\documentclass[final,3p,times]{elsarticle}
\usepackage{setspace}
\usepackage{listings}
\usepackage{xcolor}
\usepackage{graphicx}
\usepackage{float}
\usepackage{pifont}
\usepackage{caption}
\usepackage{subcaption}
\usepackage{tikz,siunitx}
\usepackage{amsmath}
\usepackage{amssymb}
\usepackage{amsthm}
\usepackage{bm}
\usepackage{soul}
\usepackage[ruled,vlined]{algorithm2e}
\usepackage[noend]{algpseudocode}

\usepackage{appendix}
\usepackage{wrapfig, blindtext}
\usepackage[utf8x]{inputenc}

\usepackage{booktabs}

\newcommand{\yawrate}[0]{{r}}

\newcommand{\vel}[0]{\boldsymbol{\nu}}
\newcommand{\mass}[0]{\mathbf{M}}
\newcommand{\coriolis}[0]{\mathbf{C}}
\newcommand{\damping}[0]{\mathbf{D}}

\newcommand{\force}[0]{\boldsymbol{\tau}}

\newcommand{\bmat}[1]{\begin{bmatrix}#1\end{bmatrix}}

\makeatletter
\newcommand{\hmat}[1]{\left[\@for\thisitem:=#1\do{\thisitem \,\,\,\,}\!\!\!\!\right]}
\makeatother

\renewcommand{\eqref}[1]{Equation (\ref{#1})}

\usepackage{comment}
\usepackage{lineno}
\usepackage{multirow}
\usepackage{adjustbox}
\usepackage{parskip} 
\usepackage[version=4]{mhchem}
\journal{Arxiv}
\theoremstyle{definition}

\title{Deep active learning for nonlinear system identification}
\definecolor{darkOrange}{rgb}{1.0,0.55,0}
\definecolor{blue}{rgb}{0.,0.,1.0}

\definecolor{defaultblue}{RGB}{31,119.,180}

\definecolor{defaultorange}{RGB}{255,168,91}

\usepackage{xcolor}
\definecolor{Tblue}{HTML}{1f77b4}
\definecolor{Torange}{HTML}{ff7f0e}
\definecolor{Tgreen}{HTML}{2ca02c}
\definecolor{Tred}{HTML}{d62728}
\definecolor{Tpurple}{HTML}{9467bd}
\definecolor{Tbrown}{HTML}{8c564b}

\usepackage{tikz}
\usepackage{pgfplots}
\pgfplotsset{compat=1.17}

\makeatletter
\newenvironment{customlegend}[1][]{%
    \begingroup
    \pgfplots@init@cleared@structures
    \pgfplotsset{#1}%
}{%
    \pgfplots@createlegend
    \endgroup
}%

\def\addlegendimage{\pgfplots@addlegendimage}
\newcommand{\todo}[1]{}

\begin{document} 
\begin{frontmatter}
\author{Erlend Torje Berg Lundby\corref{cor1}\fnref{label1}}
\ead{erlend.t.b.lundby@ntnu.no}
\cortext[cor1]{Corresponding author}
\author[label1,label2]{Adil Rasheed}
\ead{adil.rasheed@ntnu.no}
\author[label2]{Ivar Johan Halvorsen}
\ead{Ivar.J.Halvorsen@sintef.no}
\author[label1]{Dirk Reinhardt}
\ead{dirk.p.reinhardt@ntnu.no}
\author[label1]{Sebastien Gros}
\ead{sebastien.gros@ntnu.no}
\author[label1]{Jan Tommy Gravdahl}
\ead{jan.tommy.gravdahl@ntnu.no}
\address[label1]{Norwegian University of Science and Technology (NTNU), Department of Engineering Cybernetics, Trondheim, Norway}
\address[label2]{SINTEF Digital, Department of Mathematics and Cybernetics,  Trondheim, Norway}

\begin{abstract}
The exploding research interest for neural networks in modeling nonlinear dynamical systems is largely explained by the networks' capacity to model complex input-output relations directly from data. However, they typically need vast training data before they can be put to any good use. The data generation process for dynamical systems can be an expensive endeavor both in terms of time and resources. Active learning addresses this shortcoming by acquiring the most informative data, thereby reducing the need to collect enormous datasets. What makes the current work unique is integrating the deep active learning framework into nonlinear system identification. We formulate a general static deep active learning acquisition problem for nonlinear system identification. This is enabled by exploring system dynamics locally in different regions of the input space to obtain a simulated dataset covering the broader input space. This simulated dataset can be used in a static deep active learning acquisition scheme referred to as global explorations. The global exploration acquires a batch of initial states corresponding to the most informative state-action trajectories according to a batch acquisition function. The local exploration solves an optimal control problem, finding the control trajectory that maximizes some measure of information. After a batch of informative initial states is acquired, a new round of local explorations from the initial states in the batch is conducted to obtain a set of corresponding control trajectories that are to be applied on the system dynamics to get data from the system. Information measures used in the acquisition scheme are derived from the predictive variance of an ensemble of neural networks. The novel method outperforms standard data acquisition methods used for system identification of nonlinear dynamical systems in the case study performed on simulated data.

\end{abstract}

\end{frontmatter}

\section{Introduction}

Modeling dynamical systems is a cornerstone of most engineering applications where the system states of the processes change over time. Prediction models can be utilized in different ways. Models can be used in a control system setting to design safe and optimal control systems. Furthermore, if the prediction models exhibit a high degree of accuracy, they can be utilized to forecast system states over extended periods of time. Accurate simulations of the physical behavior across longer horizons can enhance our understanding of the underlying physical process and support stakeholders in informed decision making.
Physics-Based Models (PBMs) are widely utilized in a range of engineering and scientific applications. PBMs are mathematical models derived from fundamental physical principles. These principles are fundamental laws that govern a particular aspect of the natural world. Moreover, these principles or fundamental laws describe the behavior of an observable phenomenon. Our understanding of these physical phenomena is primarily attained through the examination and analysis of observed occurrences of the phenomenon in question. On one side, this means that PBMs possess a high degree of interpretability, and demonstrate generalizability when modeling assumptions are sound. On the other side, the deductive nature of this modeling approach is highly biased, and potentially ignorant to unobserved or unknown physical phenomena.
Increased access to abundant data, cheap computational resources, and many achievements and improvements in the Machine Learning (ML) society has created an enormous interest in a variety Data-Driven Models (DDMs) in many engineering fields including material science \cite{WANG2018multiscaleMaterialSci}, biomechanics \cite{feng2018ddmMicroMec}, production of biofuels \cite{peralta2012microbial}, reservoir modeling in oil and gas industry \cite{esmaili2016reservoirMod}, aluminum extraction \cite{LUNDBY2021csHAM}, bioengineering \cite{marmarelis2014data}, drug discovery \cite{gomez2018automatic} and more \cite{montans2019DDM_learn_sci_eng}.
DDMs can model an underlying process directly from input-output data. In the scientific community of systems and control, the art of building mathematical models from observed input-output data is called system identification \cite{LJUNG20101}. The research of identifying linear dynamics started in the late 1950's \cite{schoukens2019nonSysIdUserRM}. Since then, a pool of theories covering linear system identification has been developed. Linear models typically require less data and are structured and well-behaved. Thus a linear model is preferred if it can approximate a system with satisfactory accuracy. However, more complex model structures are typically required to obtain highly accurate process models of complex nonlinear systems. The inescapable nature of a wilde range of real world problems and processes is a core motivation for the broad, and ever evolving field of nonlinear system identification. Determining the model structure for nonlinear problems is a fundamental problem in nonlinear system identification. Using Neural Networks (NNs) structures stands out as highly interesting, due to their remarkable abilities to model complex nonlinear phenomena.
This has motivated researchers to investigate NNs as process models in various dynamical systems. In \cite{Naimi2020100}, NNs were used to identify the dynamics of a pressurized water nuclear reactor, authors of \cite{vargas2022NNbio} used NNs to identify the dynamics of a purification process of bioethanol, in \cite{fooshee2018DLchemRe} a Deep Learning (DL) model was used to predict chemical reactions, and in \cite{Amiruddin2020DLsysIdquad}, NNs were used to identify the dynamics of a Quadcopter. Unfortunately, NNs typically require large amounts of diverse data, which is expensive to generate and usually unavailable from dynamical systems. In \cite{Lundby2023SparseNN}, this issue is addressed by inducing sparsity in the NN models and shows that this reduces the data requirements for obtaining NNs with desirable accuracy. In addition to sparse NNs, \cite{Lundby2023SparseSkipNN} show that skip-connections can also contribute to increasing model accuracy when data size is limited. However, large amounts of training data is still needed to achieve satisfactory model accuracy. This may preclude the use of NNs in applications where data acquisition is expensive.

Active Learning (AL) aims to maximize model accuracy with a minimal amount of data by acquiring the most informative training data \cite{ren2022surveyDeepAL}. AL has been utilized in many fields, including image recognition \cite{gal2017DeepBALimage}, text classification \cite{schroder2020survey}, and object detection \cite{aghdam2019active} to mention a few. In these scenarios, large unlabeled datasets are available, and the AL algorithm aims to choose the most informative samples among the unlabeled data. The goal is then to reduce the costly labeling process which is performed by human domain experts.
In the context of dynamical systems, labeling output data may not incur significant expenses. However, data obtained from a processes under closed-loop feedback control often lacks the necessary information to identify NN models with acceptable performance, including accuracy and generalizability to operational regions of the input space, known as the state-action space.  Obtaining informative datasets of a dynamical processes is costly due to for example interruption of the production or operation, and expenses of measuring certain states. Moreover, exciting the dynamical systems to regions with high model uncertainty can induce unforeseen and potentially severe incidents to the process and its surroundings. While the safety-critical nature of using NNs in controlled processes is a major research challenge itself, approached by for example techniques within reachability analysis \cite{fan2020reachnn, manzanas2022reachability}, we limit the scope of this work to the informativeness of the sampled data, addressing challenges related to costs of sampling large datasets in dynamical systems.

Introducing AL methods to experimental design for system identification introduces additional challenges to the AL problem. That is, most AL methods address static acquisition problems where, in principle, any location in the input space is directly accessible, or a dataset is sampled in advance. For dynamical systems, on the other hand, reaching a desired location in the state-action space requires system excitation through control inputs. The topic of optimal excitation has been addressed in the research field known as optimal experiment design \cite{bombois2011optimal}. In light of this, optimal experiment design can be seen as a subfield of AL, or they can be seen as related research topics. Anyways, AL, which originates from the computer science community provides a wide range of information-theoretic approaches as well as a well-defined learning framework, providing great inspiration to researchers working with system identification.

Authors of \cite{fenet2020AL_GP_dyn} propose AL strategies for the identification of a Gaussian Process (GP) model inspired by information theoretic measures. The most promising work they propose suggests optimizing a sequence of control inputs that maximize the predictive differential entropy along the state trajectory; a method outperforming state-of-the-art experimental design methods. The work of identifying GP models was extended in \cite{tang2022AL_GP_glob_loc}, to also include global explorations. The global search for initial states is done by exploring the informativeness of short trajectories from candidate initial states. When an informative initial state is acquired, the local exploration maximizes the predictive entropy along the state trajectory as in \cite{fenet2020AL_GP_dyn}. AL is also applied to acquire data that efficiently identify linear models by solving an Optimal Control Problem (OCP) that maximize the minimal eigenvalues of covariates of states \cite{wagenmaker2020active}. An active learning approach to identify a restricted class of nonlinear dynamical modes whose state transitions depend linearly on a known feature embedding of state-action pairs was investigated in \cite{mania2020active}. However, the research on active learning for system identification of NN models is, to the best of the authors' knowledge, highly limited. 

To that end, we extend the work of AL used to acquire the most informative data for system identification to NNs. That is, 
\begin{itemize}
    \item In equation~(\ref{eq:global_exploration}), we formalize a general Batch Mode Deep Active Learning (BMDAL) acquisition scheme for dynamical system identification referred to in this work as global exploration. The scheme is on the static deep active learning batch acquisition form presented in equation~(\ref{eq:batch_DeepAL}) from \cite{ren2022surveyDeepAL}. The static acquisition scheme for dynamical systems is enabled through local explorations, obtaining a set of simulated informative candidate state-action trajectories distributed around the state-action space. The novel DeepAL scheme builds upon the AL scheme presented in \cite{tang2022AL_GP_glob_loc} that iteratively searches for the single most informative state-action trajectory for identifying a GP model through local and global explorations.
    
    \item The novel formulation of the static BMDAL acquisition for dynamical systems is utilized in a novel framework presented in Fig.~\ref{fig:DeepAL_schematic}. Ensembles of NNs are used to produce uncertainty estimates that assess the informativeness of single state-action trajectories.

    \item The general nature of the proposed BMDAL formulation allows for a wide range of query strategies to be applied. This is demonstrated by using two different query strategies in the case study.
\end{itemize}



The AL algorithm is showcased on the simulated dynamics of a 3 Degree Of Freedom (DOF) surface vessel with three states and three control inputs, yielding an input space of six dimensions. The simulator represents the dynamics of the MilliAmpere ferry\cite{Pedersen2019obs}, which is an experimental platform owed by NTNU. The simulation model is presented in Section~\ref{subsec:boat_simulator}.


%

\section{Theory}
\label{sec:Theory}
\subsection{Physics based simulator}
\label{subsec:boat_simulator}
We use the standard 3-Degrees of Freedom (3-DOF) model of a marine craft, which
is a simplified model of a real vessel. The state of the vessel is described by
the pose vector $\boldsymbol{\eta} = [x\,y\,\psi]^\top\in \mathbb{R}^3$ and the velocity vector
$\boldsymbol{\nu} = [v_1\,v_2\,r]^\top \in \mathbb{R}^3$. The pose vector describes the position and
orientation of the vessel in the North-East-Down (NED) frame with $x$ and $y$ being the position
in the North and East directions, respectively, and $\psi$ being the heading
angle. The velocity vector describes the velocity of
the vessel in the body frame with $u$ and $v$ being the velocity in the surge
and sway directions, respectively, and $r$ being the yaw rate. The model is
formulated as a nonlinear system of ordinary differential equations (ODEs) as
follows (see \cite{Fossen2011hom} for details):
\begin{equation}
    \begin{aligned}
            \dot{\boldsymbol{\eta}} &= \mathbf{R}(\psi) \boldsymbol{\nu}\\
            \mathbf{M} \dot{\boldsymbol{\nu}} + \mathbf{C}(\boldsymbol{\nu})\boldsymbol{\nu} + \mathbf{D}(\boldsymbol{\nu})\boldsymbol{\nu} &= \boldsymbol{\tau}.
    \end{aligned}
    \label{eq:simulator_boat_robot}
\end{equation}
where $\mathbf{R}(\psi) \in \mathrm{SO}(3)$ is the rotation matrix from the body
frame to the NED frame. The mass matrix $\mathbf{M} \in \mathbb{R}^{3\times 3}$,
the Coriolis matrix $\mathbf{C}(\boldsymbol{\nu}) \in \mathbb{R}^{3\times 3}$,
and the damping matrix $\mathbf{D}(\boldsymbol{\nu}) \in \mathbb{R}^{3\times 3}$
are given by \cite{Pedersen2019obs} and express the inertia and Coriolis matrices as:
\begin{equation}
    \begin{aligned}
        \mass=\bmat{
        m_{11}       & 0                               & 0            \\
        0            & m_{22}                          & m_{23}       \\
        0            & m_{32}                          & m_{33}
        },           &
        \quad
        \coriolis(\vel)=\bmat{
        0            & 0                               & c_{13}(\vel) \\
        0            & 0                               & c_{23}(\vel) \\
        c_{31}(\vel) & c_{32}(\vel)                    & 0
        },\quad
        \damping(\vel)           & =
        \bmat{
        d_{11}(\boldsymbol{\nu}) & 0                                                 & 0                        \\
        0                        & d_{22}(\boldsymbol{\nu})                          & d_{23}(\boldsymbol{\nu}) \\
        0                        & d_{32}(\boldsymbol{\nu})                          & d_{33}(\boldsymbol{\nu})
        }                                                                                                       \\
        \\
    \end{aligned}
\end{equation}
where the elements $c_{ij}(\boldsymbol{\nu})$ and $d_{ij}(\boldsymbol{\nu})$ are given by
\begin{equation}
    \label{eq:coriolis_damping}
    \begin{aligned}
        c_{13}(\vel) &= -m_{22} v_2-m_{23} \yawrate               \\
        c_{23}(\vel) &= m_{11} v_1                                \\
        c_{31}(\vel) &= -c_{13}(\vel)                                \\
        c_{32}(\vel) &= -c_{23}(\vel)
    \end{aligned}\quad
    \begin{aligned}
        d_{11}(\boldsymbol{\nu}) &=-X_{v_1}-X_{|v_1| v_1}|v_1|-X_{v_1 v_1 v_1} v_1^{2}                                         \\
        d_{22}(\boldsymbol{\nu}) &=-Y_{v_2}-Y_{|v_2| v_2}|v_2|-Y_{|r| v_2}|r|-Y_{v_2 v_2 v_2} v_2^{2}                            \\
        d_{23}(\boldsymbol{\nu}) &=-Y_{r}-Y_{|v_2| r}|v_2|-Y_{|r| r}|r|                                            \\
        d_{32}(\boldsymbol{\nu}) &=-N_{v_2}-N_{|v_2| v_2}|v_2|-N_{|r| v_2}|r|                                            \\
        d_{33}(\boldsymbol{\nu}) &=-N_{r}-N_{|v_2| r}|v_2|-N_{|r| r}|r|-N_{r r r} r^{2}
    \end{aligned}.
\end{equation}
The constant coefficients in \eqref{eq:coriolis_damping} are summarized in Table
\ref{tab:parameters}. The kinematics of the vessel is given by
\begin{equation}
    \begin{aligned}
        \dot{\boldsymbol{\eta}} &= \mathbf{R}(\psi) \mathbf{v} = \begin{bmatrix}
            \cos{\psi} & -\sin{\psi} & 0 \\
            \sin{\psi} & \cos{\psi} & 0 \\
            0 & 0 & 1
        \end{bmatrix}
        \begin{bmatrix}
            v_1 \\
            v_2 \\
            r
        \end{bmatrix} \\
    \end{aligned},
\end{equation}
and its dynamics are governed by
\begin{equation}\label{eq:kinetics}
    \dot{\vel} = \mass^{-1}\left(\force - \coriolis(\vel)\vel - \damping(\vel)\vel\right)
\end{equation}
We are interested in learning the dynamics of the vessel in response to 
given forces and moments, which we consider as control input. In the remainder
of the paper we use the following notation:
\begin{equation}\label{eq:dyn_sys_boat_simulator}
    \dot{\mathbf{x}} = \mathbf{f}(\mathbf{x}, \mathbf{u})\quad \mathrm{with}\quad \mathbf{x} = \boldsymbol{\nu}\quad \text{and}\quad \mathbf{u} = \boldsymbol{\tau},
\end{equation}
i.e. $\mathbf{x}\in\mathbb{R}^3$ is the state vector and $\mathbf{u}\in\mathbb{R}^3$ is the control input, with dynamics described by \eqref{eq:kinetics}.

\begin{table}[H]
	\centering
	\caption{Values and units of the parameters of the vessel.}
	\label{tab:parameters}
	\begin{tabular}{lll | lll}
		\toprule[1.5pt]
		{\bf Constant} & {\bf Value} & {\bf Unit} & {\bf Constant} & {\bf Value} & {\bf Unit}\\
		\midrule
		$m_{11}$ & 2389.657 &kg  & $m_{22}$ & 2533.911 &kg \\
		$m_{23}$ & 62.386 &kg    & $m_{32}$ & 28.141 &kg \\
		$m_{33}$ & 5068.910 &$\mathrm{kg \cdot m^2}$ & $X_{v_1}$ & -27.632 &$\mathrm{kg} \cdot {s}^{-1}$\\
		$X_{|v_1|v_1}$& -110.064 &$\mathrm{kg} \cdot \mathrm{s}^{-1}$ & $X_{v_1v_1v_1}$& -13.965 &$\mathrm{kg} \cdot \mathrm{s}^{-1}$ \\
		$Y_{v_2}$& -52.947 &$\mathrm{kg} \cdot \mathrm{s}^{-1}$ &	$Y_{|v_2|v_2}$ & -116.486 &$\mathrm{kg} \cdot \mathrm{s}^{-1}$\\
		$Y_{v_2v_2v_2}$& -24.313 &$\mathrm{kg} \cdot \mathrm{s}^{-1}$ &	$Y_{|r|v_2}$& -1540.383 &$\mathrm{kg} \cdot \mathrm{s}^{-1}$ \\
		$Y_r$& 24.732 &$\mathrm{kg} \cdot \mathrm{s}^{-1}$ &	$Y_{|v_2|r}$& 572.141 &$\mathrm{kg} \cdot \mathrm{s}^{-1}$\\
		$Y_{|r|r}$& -115.457 &$\mathrm{kg} \cdot \mathrm{s}^{-1}$ & $N_{v_2}$& 3.5241 &$\mathrm{kg} \cdot \mathrm{s}^{-1}$\\
		$N_{|v_2|v_2}$& -0.832 &$\mathrm{kg} \cdot \mathrm{s}^{-1}$ & $N_{|r|v_2}$& 336.827 &$\mathrm{kg} \cdot \mathrm{s}^{-1}$\\
		$N_{r}$& -122.860 &$\mathrm{kg} \cdot \mathrm{s}^{-1}$& $N_{|r|r}$& -874.428 &$\mathrm{kg} \cdot \mathrm{s}^{-1}$\\
		$N_{rrr}$& 0.000 &$\mathrm{kg} \cdot \mathrm{s}^{-1}$& $N_{|v_2|r}$& -121.957 &$\mathrm{kg} \cdot \mathrm{s}^{-1}$\\
		\bottomrule
	\end{tabular}
\end{table}

\subsection{Data-driven modeling ODE's using Neural Networks}
\begin{figure}[H]
     \centering
     \includegraphics[width=0.95\linewidth]{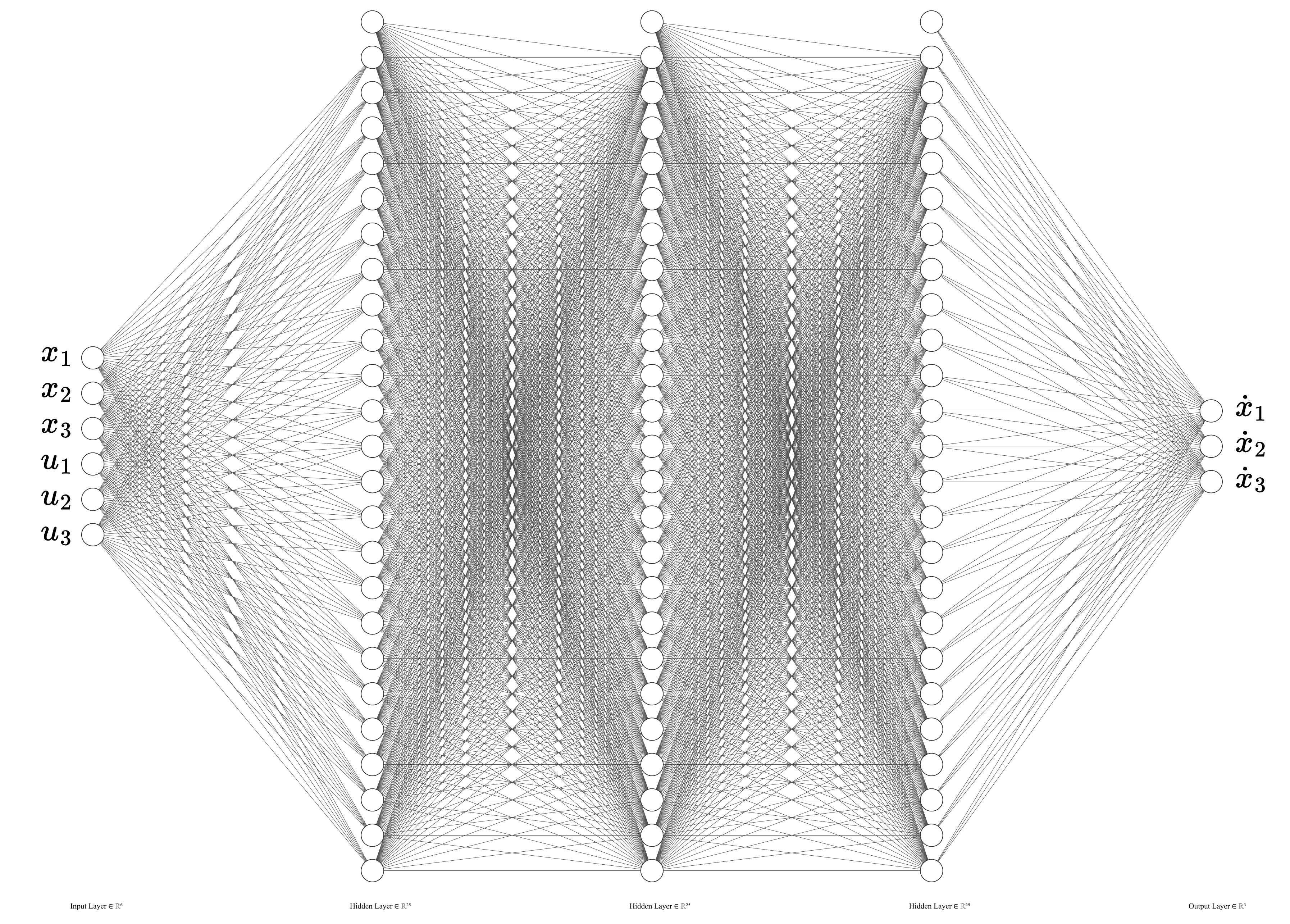}
     \caption{Neural Network architecture used}
     \label{fig:dnnarchitecture}
 \end{figure}
Data-Driven Models (DDM)s can approximate an underlying process with data from the process. The underlying process can be represented by a nonlinear mapping from input to output space. Deep feed-forward neural networks known as Multi-Layer Perceptrons (MLP) have high flexibility and, thereby ability to model nonlinear mappings. In this work, the MLP is used to model Eq. \eqref{eq:dyn_sys_boat_simulator}.
The trainable parameters of the NN are denoted by $\boldsymbol{\theta} \in \mathbb{R}^p$ where $p$ is the number of parameters, and the network is denoted as
\begin{equation}
    \Dot{\mathbf{x}} = \hat{\mathbf{f}}(\mathbf{x}, \mathbf{u}; \; \boldsymbol{\theta}),
\end{equation}
where $\Dot{\mathbf{x}}$ is the vector of time derivatives of the states. 
Optimal model parameters $\boldsymbol{\theta}^*$ are found according to the optimization of a cost function $C(\cdot)$:
\begin{equation}
    \begin{aligned}
    \boldsymbol{\theta}^* = \underset{\boldsymbol{\theta}}{\mathrm{argmin}}\left\{\frac{1}{N}\sum_{i=1}^N C(\mathbf{x}_i,\mathbf{u}_i, \Dot{\mathbf{x}}_i, \boldsymbol{\theta})   \right\}.
    \end{aligned}
    \label{eq:general_opt_NN}
\end{equation}
$N$ is the size of the training set $ {\cal D} = \{([\mathbf{x}_i,\mathbf{u}_i], \Dot{\mathbf{x}}_i)\}_{i=1}^N$, and cost function $C(\cdot)$ is given by:
\begin{equation}\label{eq:total-loss}
    C(\mathbf{x}_i,\mathbf{u}_i, \Dot{\mathbf{x}}_i, \boldsymbol{\theta}) =  J(\Dot{\mathbf{x}}_i, \hat{\mathbf{f}}(\mathbf{x}_i, \;\mathbf{u}_i; \boldsymbol{\theta})) + \lambda R(\boldsymbol{\theta}).
\end{equation} 
$J(\cdot)$ is a proper loss function. For regression tasks, the loss function is typically chosen to be the Mean Squared Error (MSE):
\begin{equation}
    J(\Dot{\mathbf{x}}_i, \hat{\mathbf{f}}(\mathbf{x}_i, \;\mathbf{u}_i; \boldsymbol{\theta})) = (\Dot{\mathbf{x}}_i - \hat{\mathbf{f}}(\mathbf{x}_i, \;\mathbf{u}_i; \boldsymbol{\theta}))^2.
\end{equation}
$R(\boldsymbol{\theta})$ is a regularization term on the parameters. In this work, the $\ell_1$ regularization is used due to its sparsity promoting nature. $\lambda$ is a tunable scalar multiplied with the regularization term $R(\boldsymbol{\theta})$. The error of the cost function is propagated backward from the output layer, through the hidden layers, and to the input layer in the so-called back-propagation scheme. 

\subsection{Ensembles of neural networks}
\label{subsec:ens_NN}
Ensemble learning includes methods that combine multiple models in making predictions. 
The main premise in ensemble learning is that errors made by individual models are likely to be compensated by other models such that the overall ensemble prediction on average improves prediction accuracy over individual models \cite{sagi2018ensemble}. Deep ensembles, consisting of multiple Deep Neural Networks (DNNs), have gained significant attention in recent years due to their improved accuracy, uncertainty estimation, and robustness to out-of-distribution data. There are two well known methods of training ensembles of NNs; by bootstrapping, where the ensemble methods are trained on different bootstrap samples of the dataset, and by multiple random seeds, where the model parameters of the members are intialized with different random parameters and then trained on the entire dataset. While the bootstrapping method has been found to hurt performance of the NNs, using random initializations turn out to be a promising approach \cite{nixon2020bootstrapped,lakshminarayanan2017simple}. 
In \cite{fort2019deep}, the success of random initialization in deep ensembles is explained by the fact that this method explore diverse modes of the function space. That is, Bayesian neural networks, which do not perform as good as deep ensembles, only explore the proximity of one single mode of the function space. Ensembles can provide both point estimates and uncertainty estimates. The point estimate can be calculated as an average of the predictions. Consider an ensemble of $M$ NNs with different parameter initialization. When forecasting several timesteps ahead without feedback from measurements, we let each ensemble member estimate individual state forecast trajectories. 
The state forecast $^j\hat{\mathbf{x}}_{t+1}$ at time $t+1$ provided by an ensemble member $\hat{\mathbf{f}}_j$ is given by forward Euler integration:
\begin{equation}
    ^j\hat{\mathbf{x}}_{t+1} = ^j\hat{\mathbf{x}}_t + 
\hat{\mathbf{f}}_j( ^j\hat{\mathbf{x}}_t,\mathbf{u}_t;\boldsymbol{\theta})\Delta T,
\label{eq:euler_forecast_ensemble_member}
\end{equation}
where $\Delta T$ is the time step. Then, the  average prediction of $M$ predictions calculated by individual NN ensemble members at timestep $t+1$ is given by:
\begin{equation}
    \boldsymbol{\mu}_{t+1} = \frac{1}{M} \sum_{j=1}^{M} \; ^{j}\hat{\mathbf{x}}_{t+1},
    \label{eq:average_forecast_ensemble}
\end{equation}
\textit{Uncertainty-based} strategies are utilized in the proposed AL method. Therefore, the main motivation of using deep ensembles in this work is to obtain an uncertainty estimate. NN predictions have two sources of uncertainty, namely model uncertainty (also known as epistemic uncertainty), and  data uncertainty (also known as aleatoric uncertainty). Usually, these types of uncertainties are modeled separately. In this study, the data is sampled from a fully observed process without process disturbances or measurement noise. Thus, only model uncertainty is considered here. The model uncertainty is caused by shortcomings in the model. This includes errors in the \textit{training procedure} such as bad training hyperparameters (learning rate, batch size, regularization etc.), \textit{insufficient model structure}, or \textit{lack of information in the data} \cite{gawlikowski2021survey}. In general, there are four different types of methods of estimating the uncertainty of a NN based on whether the NNs are deterministic or stochastic, and whether a single NN or multiple NNs are used to estimate the uncertainty.  These are \textit{single deterministic methods}, \textit{Bayesian methods}, \textit{ensemble methods} and \textit{test-time augmentation methods} \cite{gawlikowski2021survey}. 
Ensemble methods have proven to be attractive in quantifying uncertainty of NNs. Ensemble methods are in several works compared to Bayesian methods. In \cite{beluch2018power}, it is argued that ensemble based methods preform better than Bayesian Monte-Carlo Dropout approximations in DeepAL due to more calibrated predictive uncertainties. Both \cite{ovadia2019can} and \cite{gustafsson2020evaluating} came to the same conclusions, particularly under dataset shift. A simple way to quantify the uncertainty of these predictions is to calculate the empirical variance of NN predicitions for each output of the network
\begin{equation}
    \boldsymbol{\sigma}_{t+1}^2 = \frac{1}{n}\sum_{j}^n (^j\hat{\mathbf{x}}_{t+1} - \boldsymbol{\mu}_{t+1})^2.
    \label{eq:empirical_variance_pred_ens}
\end{equation}
Here, $^j\hat{\mathbf{x}}_{t+1}$ is the NN prediction made by ensemble member $j$ of $\mathbf{x}_{t+1}$, and $boldsymbol{\mu}_{t+1}$ is the mean ensemble prediction at time step $t+1$. 
\begin{figure}[H]
     \centering
    \begin{subfigure}[t]{0.45\linewidth}
        \raisebox{-\height}{\includegraphics[width=\linewidth]{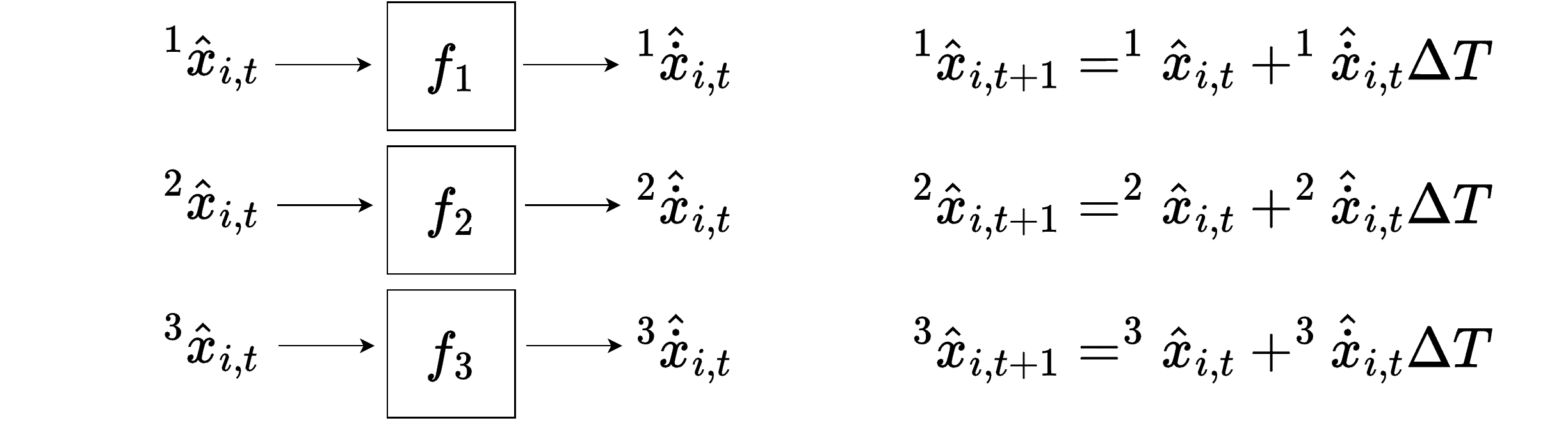}}
  \caption{GI-LI vs. GR-LR.}
        \label{subfig:ensembleEstimate}
    \end{subfigure}
    \begin{subfigure}[t]{0.45\linewidth}
        \raisebox{-\height}{\includegraphics[width=\linewidth]{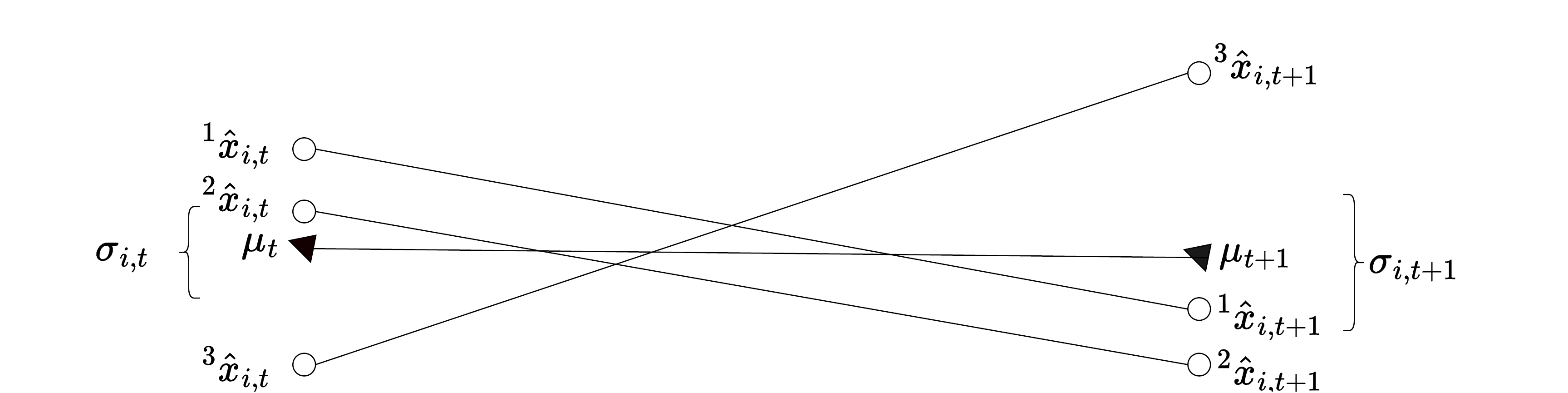}}
        \caption{GI-LI vs. GR-LI.}
        \label{subfig:enselbleEstimateIll}
    \end{subfigure}
    \caption{Ensemble}
    \label{fig:ensemble}
 \end{figure}

\subsection{Deep active learning}
Deep Active Learning (DeepAL) has emerged as a combined approach between DL and AL, addressing DL specific challenges within AL. This mainly includes dealing with over-confident uncertainty estimates of NN predictions, efficiently data-acquisition of data batches rather than the traditional AL one-by-one query method, and the joint optimization of the NN model and AL algorithm \cite{ren2022surveyDeepAL}. The majority of research on DeepAL focuses on static acquisition problems. Static acquisition problems refer to scenarios where data is already available, and any point in the input space can be acquired directly. Examples of such problems are visual data processing such as image classification \cite{wang2017costeffectiveAL} and object detection \cite{li2021DALobject}, Natural Language Processing (NLP) such as machine translation \cite{platanios2019competence}, text classification \cite{zhang2017activeDiscrTextRepLearn} and semantic analysis \cite{bhattacharjee2017ALnewsDet}. The static acquisition problem imply that there exists an unlabeled dataset $\mathcal{U} = \{\mathcal{Z}\}$ with $c$ input samples $\mathcal{Z} = \{\mathbf{z}_1, \; \mathbf{z}_2, \; ...,\; \mathbf{z}_c\}$. The goal of DeepAL in the static acquisition problem is to acquire as few as possible of the unlabeled data in $\mathcal{U}$ for labeling by choosing the most informative samples. That includes designing a query strategy $Q$, $\mathcal{U} \xrightarrow{Q}\mathcal{L}$, where $\mathcal{L} = \{ \mathcal{Z}, \mathcal{Y}\}$ is a labeled dataset, and $\mathcal{Y}$ are labels corresponding to inputs $\mathcal{Z}$. The query strategy can be expressed in terms of an acquisition function $a_{batch}$ which acquires a batch $\mathcal{B}^* = \{z_1^*, \; z_2^*, \; ..., \; z_b^*\}$ of samples to be labeled. The batch based query called BMDAL is the foundation of DeepAL. The DeepAL scheme is an iterative acquisition scheme, and one acquisition step is generally defined by:
\begin{equation}
    \begin{aligned}
        \mathcal{B}^* = \underset{\boldsymbol{\mathcal{B} \subseteq \mathcal{U} }}{\mathrm{arg max}} \quad a_{batch}(\mathcal{B},\hat{\mathbf{f}}(\mathcal{L})),
    \end{aligned}
    \label{eq:batch_DeepAL}
\end{equation}
Here $\hat{\mathbf{f}}(\mathcal{\mathcal{L}})$ is the NN. $\mathcal{L}$ is the labeled data up until the given acquisition step, and the notation $\hat{\mathbf{f}}(\mathcal{\mathcal{L}})$ indicate that the NN is trained on this data. The acquisition function $a_{batch}$ is in general a function of the NN that is trained on the currently acuired data since the informativeness of new samples can be evaluated using this NN.

The acquisition function $a_{batch}$ defines the query strategy of the AL scheme. There exists a range of different query strategies in AL. Here we will shortly describe the strategies relevant to the case study. \textit{Uncertainty-based} strategy is one of the most popular strategies in AL. The strategy aims to select samples in which the model predictions are most uncertain about. Uncertainty-based AL methods are typically computationally efficient and easy to implement. Moreover, these methods typically provides highly informative samples. One of the most utilized uncertainty-based methods calculates the predictive entropy $H[\mathbf{y}|\mathbf{x}, \mathcal{L}]$ for a given sample $\mathbf{x}$. However, there are some concerns about applying uncertainty-based sampling strategies in BMDAL. Acquiring a batch of the most informative samples using an uncertainty measure can lead to a batch of very similar samples. Moreover, strategies of this type are often focused on examples close to a decision boundary, making it vulnerable to adverserial attacks \cite{ren2022surveyDeepAL}. Hence, a \textit{Hybrid strategy} is often preferred, accounting for diversity in the sampled data. A method called Diverse Mini-Batch Active Learning (DMBAL) \cite{zhdanov2019diverse} adds informativeness to the optimization of a K-means algorithm in the weights of each candidate sample. In the DMBAL algorithm, informative estimates obtained by some informative measures are assigned as weights to the corresponding candidate samples. In each acquisition step, a batch $\mathcal{B}$ of $b$ samples closest to the centroids of the weighted K-means algorithm is added to the training set.

\subsection{Deep Active learning in Dynamical systems}
 Although data from a dynamical system may be readily available from production or operation, it often provides limited information and is not well-suited for the purpose of system identification. Due to the physical nature of dynamical systems, arbitrary points in the state-action space cannot be directly accessed. In order to sample given data points from the state-action space, the dynamics must be excited by control inputs. This is a dynamic acquisition problem. In an attempt to maximize the information contained in this sampling process, an OCP can be defined over a finite horizon, maximizing some measure of information. This is here referred to as the \textit{local exploration}. As the name indicates, this optimization is only efficient for shorter horizons, limiting the method to explore the dynamics in the proximity of the initial state. However, when identifying the input-output mapping of a nonlinear dynamical system, the entire operational window of the system must be explored. 
 
 Assuming a set of input data is already available, AL offers a robust approach for selecting the most informative data points from the input space. In the current acquisition problem for dynamical systems, we do not have access to a pre-sampled dataset. However, by locally exploring different parts of the input space, a set of simulated state-action trajectories can be obtained. With an available simulated dataset, a static AL acquisition problem for dynamical systems can be formulated. This is referred to as \textit{global exploration}. The global exploration acquires the batch of initial states corresponding to the batch of state-action trajectories that maximizes a global batch acquisition function. Following the acquisition of a set of initial states through global exploration, a subsequent round of local exploration is conducted for each state in the batch. This local exploration entails a longer optimization horizon compared to the initial search conducted for all candidates during the global exploration. This is because the computational complexity of the OCP increases significantly with the horizon, making it necessary to restrict the horizon to a relatively short length when optimizing for all candidates prior to the global exploration. When control trajectories are obtained from the final local explorations these trajectories are applied to the real system from the corresponding acquired initial states. This is done under the assumption that the system is driven to each initial state using a specific control law. As the system evolves under the applied control sequences, data on the system states is collected.
 
\subsubsection{Local exploration}
Data sampled from a dynamical system should be properly excited by a control signal to obtain informative data that can be used for system identification. Local exploration refers to the dynamic AL acquisition problem of finding a control trajectory that informatively excites the system. Given an initial state $\mathbf{x}_0$ from where the dynamical system is excited, the local exploration can be formulated as an open loop finite horizon OCP , which yields a sequence of control inputs $\{\mathbf{u}\}_{k=0}^{T-1, \;*} = \{\mathbf{u}_0, \; \mathbf{u}_1,\; ...,\; \mathbf{u}_{T-1} \}^*$ . In the context of active learning, the objective function is an acquisition function $a_{local}$ that measures the informativeness of the sequence of forecasted states $\{\hat{\mathbf{x}}_t\}_{t=1}^T = \{ \hat{\mathbf{x}}_1, \; \hat{\mathbf{x}}_2, \; ...,\; \hat{\mathbf{x}}_T\}$ given the candidate sequence of control inputs $\{\mathbf{u}_{k}\}_{k=0}^{T-1} = \{ \mathbf{u}_0,\; \mathbf{u}_1,\; ..., \; \mathbf{u}_{T-1} \}$ and an initial state $\mathbf{x}_0$:
\begin{equation}
    \begin{aligned}
    \{\mathbf{u}_t\}_{t=0}^{T-1, \; *} &= \underset{\{\mathbf{u}_{t}\}_{t=0}^{T-1}}{\mathrm{arg max}} \; a_{local}\left( \mathbf{x}_0, \; \{\mathbf{u}_{t}\}_{t=0}^{T-1}, \{\hat{\mathbf{x}}_t\}_{t=1}^{T}\right) \\
    s.t. \quad \hat{\mathbf{x}}_{t+1} &= \hat{\mathbf{f}}(\hat{\mathbf{x}}_{t}, \mathbf{u}_t;\boldsymbol{\theta})
    \end{aligned}
    \label{eq:local_exploration}
\end{equation}
where $\hat{\mathbf{x}}_{0} = \mathbf{x}_{0}$. The standard strategy in Model Predictive Control (MPC) formulation is to only apply the first control input in the sequence and then solve the OCP again for each consecutive timestep until the end of the horizon. This scheme requires $T-1$ optimizations to obtain one sequence of control inputs, and is therefore computational expensive. An alternative that is computationally feasible is to optimize for the entire control sequence one time and apply the control sequence obtained from that one solution of the OCP. The authors of \cite{fenet2020AL_GP_dyn} developed an active learning scheme for a GP model. They suggested to maximize the sum of differential entropy of the GP model predictions over the control horizon of $T$ steps, such that $a_{local} = \sum_{i=0}^{T-1} H[\hat{\mathbf{f}}(\hat{\mathbf{x}}_i,\mathbf{u}_i; \boldsymbol{\theta})]$.
The differential entropy of variable $y$ is defined by \cite{cover1999elements}
\begin{equation}
    H(y) = -\int_{\mathcal{Y}}p(y) log(p(y)) dy,
\end{equation}
where $p(y)$ is the probability density function. In this case, the probability density function represent the distribution over the predictions. If the variable $y$ is Gaussian distributed, the differential entropy is given by
\begin{equation}
    H_{Gaussian}(y) = \frac{1}{2}log(2\pi exp(\sigma^2(y)),
\end{equation}
where $\sigma^2$ is the variance of the given prediction. 
\subsubsection{Global exploration}
Exciting the system dynamics is essential to obtain informative data from a dynamical system. The local exploration formulated as an OCP in equation~(\ref{eq:local_exploration}) provides a sound basis for exiting the problem locally. However, when the goal is to obtain the most informative data from the entire input space, solely depending on the optimization in equation~(\ref{eq:local_exploration}) is inefficient. That is, the computational complexity increases drastically with optimization horizon. This put restrictions on how long the optimization horizon can be, and therefore also the area that a optimized state-action trajectory can span. Moreover, the uncertainty of state forecasts typically increases with each time step. This is highly relevant in the local exploration formulation since the corresponding OCP typically aims to maximize some uncertainty measure. With high levels of uncertainty, the actual states are likely to deviate from predicted states after longer horizons. Thus, the efficacy of local explorations as defined above is typically limited to exploring dynamics locally. Authors in \cite{tang2022AL_GP_glob_loc} suggest partitioning the search problem into global and local explorations for actively learning a GP model. Building upon the work in \cite{tang2022AL_GP_glob_loc}, equation~(\ref{eq:global_exploration}) provides a general formulation of the DeepAL optimization problem for dynamical systems, acquiring an optimal batch rather than single initial states. The global exploration consider a set $\mathcal{X} = \{\mathbf{x}_{0, \; 1}, \; \mathbf{x}_{0, \; 2}, \; ...,\; \mathbf{x}_{0, \; c}\}$ of $c$ candidate initial states. For each of the candidate initial states $\mathbf{x}_{0, \; i}$, an optimal control trajectory $(\mathbf{u_t}_{t=0}^{T-1, \;*})_i$ is obtained by optimizing the OCP in equation~(\ref{eq:local_exploration}). With an initial condition $\mathbf{x}_{0, \; i}$ and the obtained control trajectory $(\mathbf{u_t}_{t=0}^{T-1, \;*})_i$, the corresponding forecasted state trajectory $\left(\{\hat{\mathbf{x}}_t\}_{t=1}^{T} \right)_i$ is estimated by the model. One acquisition step of the global exploration is generally described in the following DeepAL optimization formulation:
\begin{equation}
\begin{aligned}
    \mathcal{B}^* &= \underset{\mathcal{B \subseteq \mathcal{X}}}{argmax} \; a_{global}\left(\mathcal{B}, \; \left \{\{\mathbf{u}_{t}\}_{t=0}^{T-1,\; *}, \{\hat{\mathbf{x}}_t\}_{t=1}^{T} \right\}_{i=1}^b \right) \\
    & \left \{\{\mathbf{u}_{t}\}_{t=0}^{T-1,\; *}, \{\hat{\mathbf{x}}_t\}_{t=1}^{T} \right\}_{i=1}^b \subseteq \mathcal{XU}
    \end{aligned}
    \label{eq:global_exploration}
\end{equation}
where $\mathcal{B} = \{\mathbf{x}_{0,\;1}, \; \mathbf{x}_{0,\;2}, \; ...,\; \mathbf{x}_{0,\;b} \}$ is a candidate batch of initial conditions, and $\left \{\{\mathbf{u}_{t}\}_{t=0}^{T-1,\; *}, \{\hat{\mathbf{x}}_t\}_{t=1}^{T} \right\}_{i=1}^b$ is the corresponding batch of simulated state-action trajectories. $\mathcal{XU} = \left\{\left\{\mathbf{u}_{t}\}_{t=0}^{T-1,\; *}, \{\hat{\mathbf{x}}_t\}_{t=1}^{T} \right\}_1, \; \left\{\mathbf{u}_{t}\}_{t=0}^{T-1,\; *}, \{\hat{\mathbf{x}}_t\}_{t=1}^{T} \right\}_2, \; ..., \; \left\{\mathbf{u}_{t}\}_{t=0}^{T-1,\; *}, \{\hat{\mathbf{x}}_t\}_{t=1}^{T} \right\}_c  \right\}$ is the set of simulated candidate state-action trajectories.  The acquired a batch $\mathcal{B}^* = \{\mathbf{x}_{0,\;1}^*, \; \mathbf{x}_{0,\;2}^*, \; ..., \; \mathbf{x}_{0,\;b}^*\}$ of initial conditions corresponds to the batch simulated state-action trajectories $\left (\{\mathbf{u}_{t}\}_{t=0}^{T-1,\; *}, \{\hat{\mathbf{x}}_t\}_{t=1}^{T} \right)_{i=1}^b$ that maximize some global batch acquisition function $a_{global}$. Since simulated state-action trajectories are already sampled from in a local exploration scheme, the global exploration becomes a static acquisition problem on the form of the standard DeepAL scheme presented in equation~(\ref{eq:batch_DeepAL}).  

\section{Method and setup}
 In this section, the experimental setup, as well as the methods used in the case study is presented. The data is generated by integrating the nonlinear ODEs in equation~(\ref{eq:dyn_sys_boat_simulator}) with a set of initial values for the states $\mathbf{x}_0$ using the fourth-order Runge-Kutta (RK4) numerical integration algorithm. In the DeepAL method presented in this work, a batch of initial states are chosen from a set of candidate states according to the optimization in equation~(\ref{eq:global_exploration}). The query strategies defined by the global acquisition function $a_{global}$ are described in Section~\ref{subsec:IntegratedDeepAL}. The control trajectories that excite the system dynamics are acquired in the local exploration scheme defined in equation~(\ref{eq:local_exploration}). The local acquisition function $a_{local}$ in this scheme is an \textit{uncertainty-based strategy} also described in detail in Section~\ref{subsec:IntegratedDeepAL}. The benchmark method chooses the set of initial conditions randomly. Moreover, the control input trajectory from each initial state are chosen according to the Amplitude modulated Pseudo Random Binary Signal (APRBS) used to identify nonlinear dynamics with NNs in works like \cite{WINTER2018802}, \cite{Lundby2023SparseSkipNN} and \cite{robinson2022novel}.
 In each loop of the AL scheme, a batch of $b=10$ time series $\{\mathbf{X}_1, \; \mathbf{X}_2, \; ..., \; \mathbf{X}_i, \; ..., \; \mathbf{X}_{b} \}$ are obtained. The $i'th$ time series is obtained by simulating the dynamics from an initial condition $\mathbf{x}_0 = [x_1(0), \; x_2(0),\; x_3(0)]$ over a horizon of $T=15$ with timesteps $\Delta T = 0.1 sec$. This yields a time series $\mathbf{X}_i$:
 \begin{equation}
     \mathbf{X}_i = 
     \begin{bmatrix}
     x_1(0) & x_2(0) & x_3(0) & u_1(0) & u_2(0) & u_3(0) \\
     x_1(1) & x_2(1) & x_3(1) & u_1(1) & u_2(1) & u_3(1) \\
     \vdots & \vdots & \vdots & \vdots & \vdots & \vdots \\
     x_1(t) & x_2(t) & x_3(t) & u_1(t) & u_2(t) & u_3(t) \\
     \vdots & \vdots & \vdots & \vdots & \vdots & \vdots \\
     x_1(T-1) & x_2(T-1) & x_3(T-1) & u_1(T-1) & u_2(T-1) & u_3(T-1) \\
     x_1(T) & x_2(T) & x_3(T) & nan & nan & nan
     \end{bmatrix}
 \end{equation}
Hence, the control inputs $\mathbf{u}(t)$  are defined until timestep $t = T-1$. That is, at the last step there is no need for a control input since there is no next state to be calculated. For each time series $i$ in the batch, the output label $\mathbf{Y}_i$ for training is calculated by the forward Euler formula:
\begin{equation}
    \mathbf{Y}_i = 
    \begin{bmatrix}
        \frac{x_1(1) - x_1(0)}{\Delta T} & \frac{x_2(1) - x_2(0)}{\Delta T} & \frac{x_3(1) - x_3(0)}{\Delta T}\\
        \frac{x_1(2) - x_1(1)}{\Delta T} & \frac{x_2(2) - x_2(1)}{\Delta T} & \frac{x_3(2) - x_3(1)}{\Delta T}\\
        \vdots & \vdots & \vdots \\
        \frac{x_1(T) - x_1(T-1)}{\Delta T} & \frac{x_2(T) - x_2(T-1)}{\Delta T} & \frac{x_3(T) - x_3(T-1)}{\Delta T}\\
    \end{bmatrix}
\end{equation}
We define a new matrix $\mathbf{X}_i^{'}$ that contains all but the last row of the $i'th$ time series $\mathbf{X}_i$. Then $\mathbf{X}_i^{'}$ and $\mathbf{Y}_i$ are paired as inputs and outputs:
\begin{equation}
    \mathcal{S}_i = [\mathbf{X}_i^{'}, \; \mathbf{Y}_i].
\end{equation}
This is done for all simulations in the batch. Then the input-output pairs are stacked:
\begin{equation}
    \mathcal{S}_{batch} = [\mathcal{S}_1^T, \; \mathcal{S}_2^T,\; ..., \; \mathcal{S}_b^T]^T,
\end{equation}
before added to the training data $\mathcal{D}_{train}$. Before the training is conducted, the inputs and outputs in the training set $\mathcal{D}_{train}$ are normalized, shuffled, and put in mini-batches for training. In each loop of the learning scheme, an ensemble of $M=10$ NNs are trained on all training data sampled up until that time.

\subsection{Novel DeepAL scheeme for dynamical systems}
\label{subsec:IntegratedDeepAL}
The DeepAL acquisition scheme comprises a global exploration scheme and a local exploration scheme. The global exploration scheme will for each acquisition step in the AL loop choose a batch of $b$ initial states $\mathcal{B}^* = \{\mathbf{x}_{0,\;1}^*, \; \mathbf{x}_{0,\;2}^*, \; ..., \; \mathbf{x}_{0,\;b}^* \}$ among a set of $c$ candidates $\mathcal{X} = \{\mathbf{x}_{0,\;1}, \; \mathbf{x}_{0,\;2}, \; ..., \; \mathbf{x}_{0,\;c} \}$ according to the AL optimization problem defined in equation~(\ref{eq:global_exploration}). For each of the candidate initial states $\mathbf{x}_{0,\; i}$ a state-action trajectory is obtained according to the local exploration in equation~(\ref{eq:local_exploration}). The query strategy is defined by the global batch acquisition function $a_{global}$, which quantifies the informativeness of batches of state-action trajectories corresponding to initial state candidates. A simple \textit{uncertainty}-based acquisition function that sum of predictive entropies along all candidate trajectories is given by:
\begin{equation}
    a_{global}\left(\mathcal{B}, \left(\{\mathbf{u}_{t}\}_{t=0}^{T-1,\; *}, \{\hat{\mathbf{x}}_t\}_{t=1}^{T} \right)_{i=1}^b\right) = \sum_{i=1}^b\sum_{t=1}^T H([^j\hat{\mathbf{x}}_t]_{j=1}^M),
\end{equation}
where $[^j\hat{\mathbf{x}}_t]_{j=1}^M$ is the set of ensemble forecasts at timestep $t$. Assuming that the ensemble predictions are approximately normally distributed around the mean prediction $\boldsymbol{\mu}_t$ given in equation~(\ref{eq:average_forecast_ensemble}), and that predicted states are  uncorrelated, maximizing the entropy will become approximately the same as to maximizing the empirical variance given in equation~(\ref{eq:empirical_variance_pred_ens}):
\begin{equation}
    H([^j\hat{\mathbf{x}}_t]_{j=1}^M) \approx \boldsymbol{\sigma}_{t}^2.
\end{equation}
In order to scale the optimization problem according to the magnitude of states in the state vector, the empirical variance of state $k$, $\sigma_{t, \;k }$, $k \in \{1,\; 2, \; 3\}$ can be divided by the standard deviation of the $k'th$ state based on the currently sampled dataset. defining the vector $\mathbf{s}_{inv} = [\frac{1}{std_1}, \; \frac{1}{std_2}, \; \frac{1}{std_3}]^T$, where $std_k$ is the standard deviation of the $k'th$ state $x_k$, the resulting acquisition function can be defined as:
\begin{equation}
    a_{global}\left(\mathcal{B}, \left(\{\mathbf{u}_{t}\}_{t=0}^{T-1,\; *}, \{\hat{\mathbf{x}}_t\}_{t=1}^{T} \right)_{i=1}^b\right) = \sum_{i=1}^b\sum_{t=1}^T \boldsymbol{\sigma}_{t}^2 \odot \mathbf{s}_{inv},
    \label{eq:uncertainty_acqusition_global}
\end{equation}
where $\odot$ is the Hadamard product operator that takes the element-wise multiplication of the two vectors. The resulting acquisition function is purely uncertainty based and does not take into account the similarity between samples. Hybrid acquisition strategies takes into account both uncertainty of individual samples as well as the similarities between samples in a candidate batch $\mathcal{B}$. An intuitive hybrid acquisition method DMBAL adds informativeness to the optimization of a weighted K-means algorithm. That is, the algorithm acquires the closest sample to each of the $b$ centroids found by a weighted K-means, where the weight is some informative measure. A simple adaption of the algorithm to the problem at hand is given in Algorithm~\ref{alg:diverse_mini_batch_active_learning}:
\begin{algorithm}
\caption{Diverse Mini-Batch Active Learning (DMBAL) in Global exploration}
\label{alg:diverse_mini_batch_active_learning}
\textbf{Input:} Candidate initial conditions $\mathcal{X}$, Acquired dataset  $\mathcal{D}_{train}$, \\
pre-filter factor $\beta$, batch size/number of clusters $b$, Required level of model accuracy $\alpha$ \\
Train model on $\mathcal{D}_{train}$\\
\While{required level of model accuracy $\alpha$ is not reached}{
    Get informativeness $\sum_{t=1}^T \boldsymbol{\sigma}_{t}^2 \odot \mathbf{s}_{inv}$ for simulated state-action trajectories corresponding\\
    to initial states in $\mathcal{X}$ \\
    Prefilter top $\beta \cdot b$ informative samples \\
    Cluster $\beta \cdot b$ initial states to $b$ clusters with weighted K-means \\
    Select batch $\mathcal{B}^*$ of $b$ different initial states closest to the cluster centers\\
    Perform local exploration to obtain control input trajectories for each initial state in $\mathcal{B}^*$\\
    From initial conditions in $\mathcal{B}^*$, apply obtained control trajectories on system dynamics \\
    Add sampled data to training set $\mathcal{D}_{train}$\\
    Train model on all samples in $\mathcal{D}_{train}$
}
\end{algorithm}
The method reuses the uncertainty measure defined in equation~(\ref{eq:uncertainty_acqusition_global}). In addition, the method aims to add diversity of samples by comparing the similarities of the candidate initial conditions in the modified DMBAL method. The $c$ candidates in $\mathcal{X} = \{ \mathbf{x}_{0,\;1}, \mathbf{x}_{0,\; 2}, \; ..., \mathbf{x}_{0, c} \}$ are at each acquisition step uniformly sampled from the intervals given in Table~\ref{table:init_condition_candidates}: 
\begin{table}[H]
        \begin{center}\caption{Initial condition intervals for the states $\mathbf{x}$}
        \begin{tabular}[width=\textwidth]{l|l}
        \hline
        Variable &  Initial condition interval \\ \hline
        $x_1$ & $[-0.2,\;1.4]$\\
        $x_2$ & $[-0.2,\; 0.2]$ \\
        $x_3$ & $[-0.2,\; 0.2]$  \\
        \hline
        \end{tabular}
        \label{table:init_condition_candidates}
        \end{center}
        \end{table}

\begin{figure}[H]
     \centering
     \includegraphics[width=0.95\linewidth]{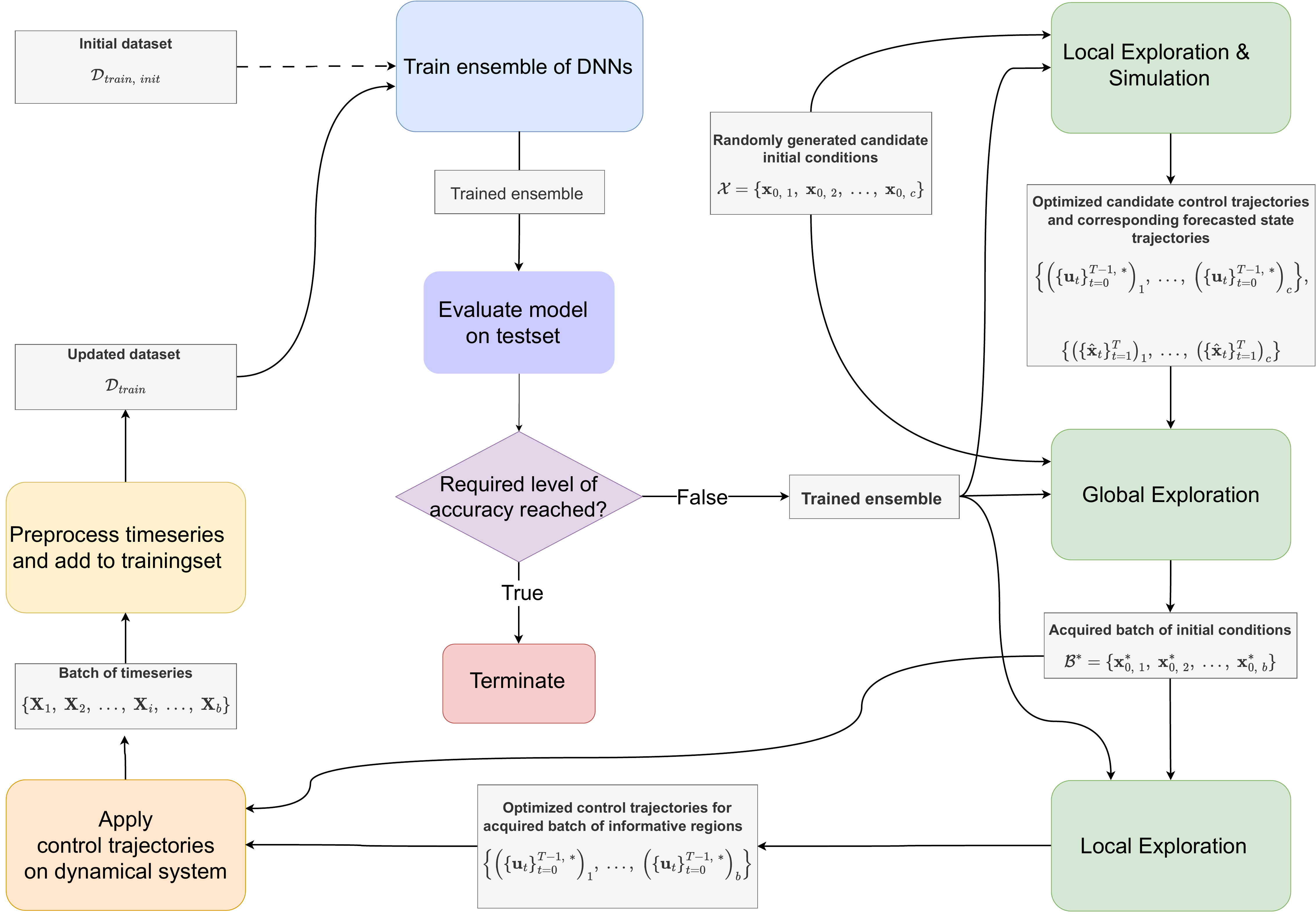}
     \caption{Schematic presentation of the DeepAL scheme. Given an initial dataset $\mathcal{D}_{init}$, NNs in an ensemble are trained. If the ensemble yields the required level of model accuracy, the AL scheme is terminated. If not, the ensemble is used in global and local explorations. First, local exploration and simulation procedure generates control trajectories for each initial state candidate in $\mathcal{X}$. The global exploration scheme then acquires a batch $\mathcal{B}^*$ of $b$ initial states in which excitation of the dynamics would be most informative. Then, local explorations are conducted for each of the initial states in $\mathcal{B}^*$, yielding a set of $b$ control trajectories. Given that the dynamics are driven to each of the initial conditions in $\mathcal{B}^*$, the corresponding control trajectories are applied to excite the dynamics. From this excitation process, time series are obtained, preprocessed and added to the training set $\mathcal{D}_{train}$. The ensemble is then trained on this training set, and the procedure is repeated until the required model accuracy on the test set is achieved, or a sampling budget is exhausted.}
     \label{fig:DeepAL_schematic}
 \end{figure}

The local exploration scheme obtains a sequence of control inputs by optimizing the dynamic acquisition problem formulated as an OCP in equation~(\ref{eq:local_exploration}) from a given initial state. The local acquisition function $a_{local}$ used in the optimization defined in equation~(\ref{eq:local_exploration}) is the same as the uncertainty-based global acquisition function defined in equation(~\ref{eq:uncertainty_acqusition_global}), but only for a single initial state and the corresponding trajectory. That is, the local acquisition function is:
\begin{equation}
    a_{local} = \sum_{t=1}^T \boldsymbol{\sigma}_{t}^2 \odot \mathbf{s}_{inv}.
\end{equation}
A schematic illustration of the novel DeepAL scheme is presented in Fig.~\ref{fig:DeepAL_schematic} 
The OCP solved to generate control sequence is solved in the optimization framework CasADi \cite{Andersson2019Casadi}, while the design and training of DNNs used in the optimization is done using the DL framework PyTorch \cite{paszke2019pytorch}. The ML-CasADi package developed by authors of \cite{salzmann2023mlCasadi} is used to integrate the two frameworks.
\subsection{Performance metrics}
We focus on the forecast error for several steps ahead in a so called \textit{rolling forecast} to measure the generalization error of the NNs. That is, given initial conditions $\mathbf{x}_0 = \mathbf{x}(t_0)$ and a sequence of $n$ control inputs $\{\mathbf{u}(t_0),\; \mathbf{u}(t_1),\; ...,\;\mathbf{u}(t_{n-1})\}$,  a NN forecast the consecutive $n$ timesteps $\{\hat{\mathbf{x}}(t_1), \; \hat{\mathbf{x}}(t_2), \; ..., \; \hat{\mathbf{x}}(t_n) \}$. The NN $\hat{\mathbf{f}}$ predict the time derivatives of the states at timestep $i$, that is $d\mathbf{x}(t_i)/dt$, based on the current state prediction $\hat{\mathbf{x}}(t_i)$ and control input $\mathbf{u}(t_i)$:
\begin{equation}
    \frac{d \hat{\mathbf{x}}(t_i)}{dt} = 
    \begin{cases}
    \hat{\mathbf{f}}(\hat{\mathbf{x}}(t_i), \mathbf{u}(t_i)), & \text{if } \mathbf{x}(t_i) \text{ is not measured,}\\
    \hat{\mathbf{f}}(\mathbf{x}(t_i), \mathbf{u}(t_i)), & \text{if } \mathbf{x}(t_i) \text{ is measured.}
    \end{cases}    
\end{equation}
Hence, the model either uses the previous calculated forecast of the state $\hat{\mathbf{x}}(t_i)$ or the true state $\mathbf{x}(t_i)$ depending on whether the state is measured or not. Here, we assume no measurement noise, making the measured and true state the same. In the case study, the states are measured at every $50$ timestep, meaning that the model forecast for this horizon before it is corrected by measurements. After estimating the time derivative at timestep $i$, the forecast at the next timestep is calculated by the forward Euler formula:
\begin{equation}
    \hat{\mathbf{x}}(t_{i+1}) = \hat{\mathbf{x}}(t_i) + \frac{d \hat{\mathbf{x}}(t_i)}{dt} \cdot \Delta T.
\end{equation}
The rolling forecast accuracy measure used in for example \cite{robinson2022novel}, \cite{Lundby2023SparseNN} and \cite{Lundby2023SparseSkipNN}  called Average Normalized Rolling Forecast Mean Squared Error (AN-RFMSE) is a scalar defined by
\begin{equation}
    \textrm{AN-RFMSE} = \frac{1}{p}\sum_{i=1}^p\frac{1}{n}\sum_{j=1}^n\left(\frac{\hat{x}_i(t_j) - x_i(t_j)}{std(x_i)}\right)^2,
    \label{eq:AN-RFMSE}
\end{equation}
where $\hat{x}_i(t_j)$ is the model estimate of the simulated state variable $x_i$ at time step $t_j$, $std(x_i)$ is the standard deviation of variable $x_i$ in the training set $\mathcal{D}_{train}$, $p=3$ is the number of state variables and $n$ is the number of time steps the normalized rolling forecast MSE is averaged over. 
\subsection{Test set generation}
The utilization of DNNs in modeling dynamical systems is driven by their capability to represent intricate relationships with a high degree of accuracy. When proper measures are taken to address safety considerations, they have the potential to enhance the optimality of MPC. As a result, evaluating the sampling strategies on a testset generated by using an optimal control policy is a subject of significant interest. 
The MPC used when generating the testset solves an OCP that minimize a quadratic cost function:
\begin{equation}
\begin{aligned}
 \{\mathbf{u}_0, \; ...,\; \mathbf{u}_{n-1} \}   &= \underset{\mathbf{u}}{\mathrm{min}} 
 \sum_{k=0}^{n-1} (\mathbf{x}_k - \mathbf{x}_{ref,\;k})^T\mathbf{Q}(\mathbf{x}_k - \mathbf{x}_{ref, \; k}) + \mathbf{u}_k\mathbf{R}\mathbf{u}_k,
 \label{eq:mpc_quadratic}\\
 \text{s.t.} \quad \dot{\mathbf{x}}_k &= \mathbf{f}(\mathbf{x}_k, \mathbf{u}_k).
 \end{aligned}
\end{equation}
$\mathbf{x}_{ref,\;k}$ is the desired reference signal, and $\mathbf{Q}$ and $\mathbf{R}$ are weighting matrices. The subscript $k$ indicates the value of the variable at timestep $k$. The function $\mathbf{f}(\cdot, \cdot)$ is the simulation model itself. Given an initial condition, the optimization problem in equation~\eqref{eq:mpc_quadratic} is solved for $n$ steps. Both the sequence of states $\{\mathbf{x}_k\}_{k=0}^{n-1}$ and control inputs $\{\mathbf{u}_k\}_{k=0}^{n-1}$ are decision variables in the optimization, and can be extracted from the solution of the optimization. The testset consist of $100$ time series with initial conditions uniformly sampled from the intervals in Table~\ref{table:init_condition_candidates}.
\begin{table}[H]
        \begin{center}\caption{Reference intervals for the states $\mathbf{x}_{ref}$}
        \begin{tabular}[width=\textwidth]{l|l}
        \hline
        Variable &  Reference values interval \\ \hline
        $x_1$ & $[-0.3,\;1.3]$\\
        $x_2$ & $[-0.3,\; 0.3]$ \\
        $x_3$ & $[-0.3,\; 0.3]$  \\
        \hline
        \end{tabular}
        \label{table:reference_intervals}
        \end{center}
        \end{table}
Each of the time series are generated by ten optimizations of equation~\eqref{eq:mpc_quadratic} each with a horizon of $50$ timesteps. The final state $\mathbf{x}_{n}$ in one optimization is then the initial state of the next time series, such that the $i'th$ time series in the testset can be written as:
\begin{equation}
    \mathbf{X}_{test,\ i} = 
    \begin{bmatrix}
        x_1(0) & \hdots & u_3(0) \\
        \vdots & \ddots & \vdots \\
        x_1(n-1) & \hdots & u_3(n-1)\\
        x_1(n) & \hdots & u_3(n) \\
        \vdots & \ddots & \vdots \\
        x_1(10*n - 1) & \hdots & u_3(10*n -1)
    \end{bmatrix},
\end{equation}
where $n=50$. The value of the references $\mathbf{x}_{ref}$ are uniformly drawn from Table~\ref{table:reference_intervals} are constant for the optimization horizon. Hence each time series has ten different references over $500$ timesteps. Hence, the testset can be written as:
\begin{equation}
    \mathcal{D}_{test} = \left\{X_{test, \;1}, \; X_{test, \;2}, \; ...,\; X_{test, \;100}   \right\}
\end{equation}
 
\section{Results and discussion}
\label{sec:resultsanddiscussion}
The case study presented in this section investigate the efficacy of global and local explorations compared to benchmark random sampling methods. Moreover, the study presents the effect of the global hybrid strategy DMBAL for different values of the prefilter hyperparameter $\beta$, where the special case of $\beta = 1$ can be considered as an uncertainty based acquisition strategy. 
\subsection{Information based and random sampling}
\begin{figure}[H]
    \centering
    \includegraphics[width=0.8\linewidth]{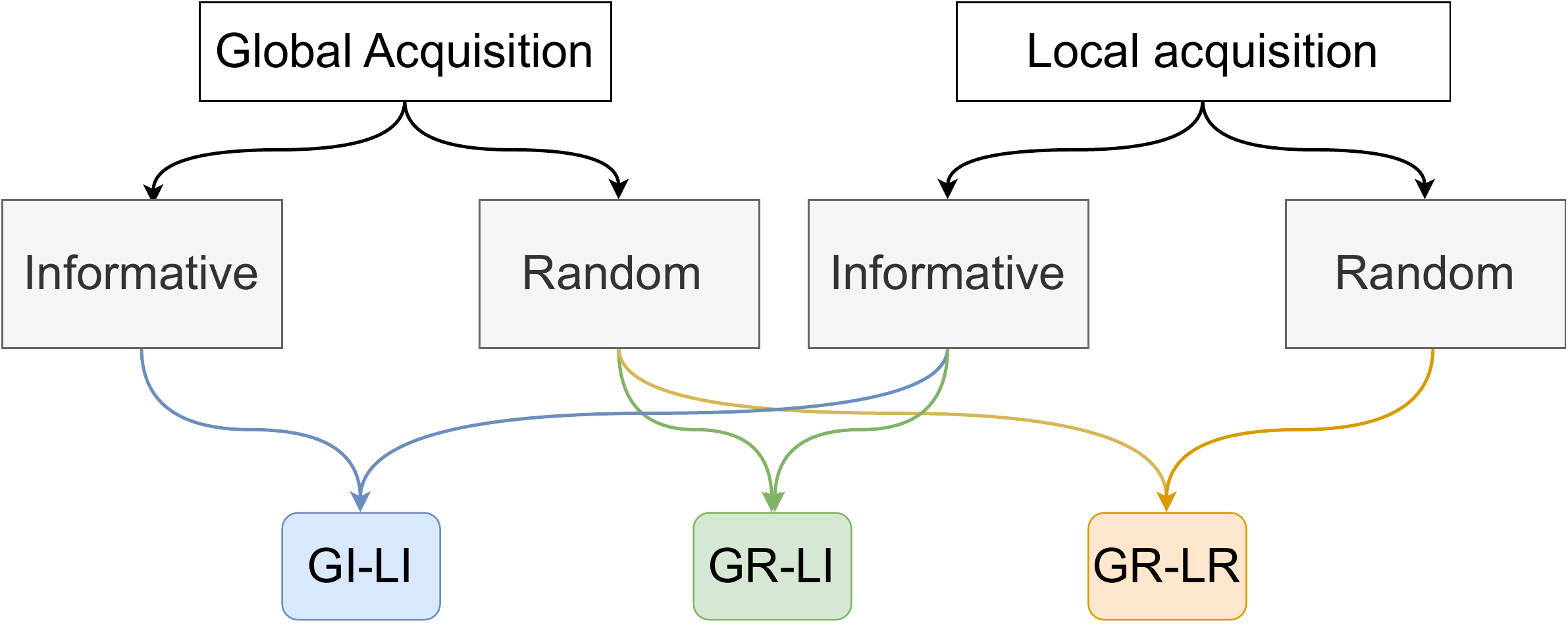}
    \begin{adjustbox}{max width=1\linewidth}
        \begin{tikzpicture}
            \begin{customlegend}[legend columns=6, legend style={draw=none, column sep=2ex}, legend entries={GI-LI ,GR-LI ,GR-LR}]
                \addlegendimage{Tblue!40, fill=Tblue!20, area legend}
                \addlegendimage{Tgreen!40, fill=Tgreen!20, area legend}
                \addlegendimage{Torange!40, fill=Torange!20, area legend}
            \end{customlegend}
        \end{tikzpicture}
    \end{adjustbox}
    \caption{Schematic of how fundamental sampling schemes are derived from global and local sampling strategy. Global Informative Local Informative(GI-LI), Global Random Local Informative(GR-LI), and Global Random Local Random(GR-LR) are the derived methods that are investigated in the case study.}
    \label{fig:global_local_info_random}
\end{figure}

  In order to quantify the efficacy of global and local explorations in the proposed DeepAL sampling scheme compared to benchmark random sampling methods, we define three fundamental data acquisition schemes.  All schemes use both a global-, and a local sampling method. The three fundamental methods are based on either an information theoretic approach or a random sampling strategy for both local and global exploration. Fig.~\ref{fig:global_local_info_random} shows a schematic presentation of how the different schemes are derived from the two sampling strategies.
The fundamental sampling schemes are namely Global Informative Local Informative(GI-LI), Global Random Local Informative(GR-LI), and Global Random Local Random(GR-LR). The  GI-LI method is described in Section~\ref{subsec:IntegratedDeepAL}, and is using information theoretic sampling strategies both locally and globally. The global exploration in GI-LI acquires initial conditions from where to conduct the local explorations. The global exploration is using the local exploration method to assign measures of informativeness to the candidate initial conditions. The GR-LI method is globally random, and a batch of initial states is acquired by uniformly sampling within the interval of states presented in Table~\ref{table:init_condition_candidates}. The local exploration method of GR-LI is the exact same as the local sampling method used GI-LI. Th GR-LR method uses the same random global strategy as GR-LI, and uses the APRBS method to excite the dynamics locally. 
\begin{figure}[H]
     \centering
    \begin{subfigure}[t]{0.45\linewidth}
        \raisebox{-\height}{\includegraphics[width=\linewidth]{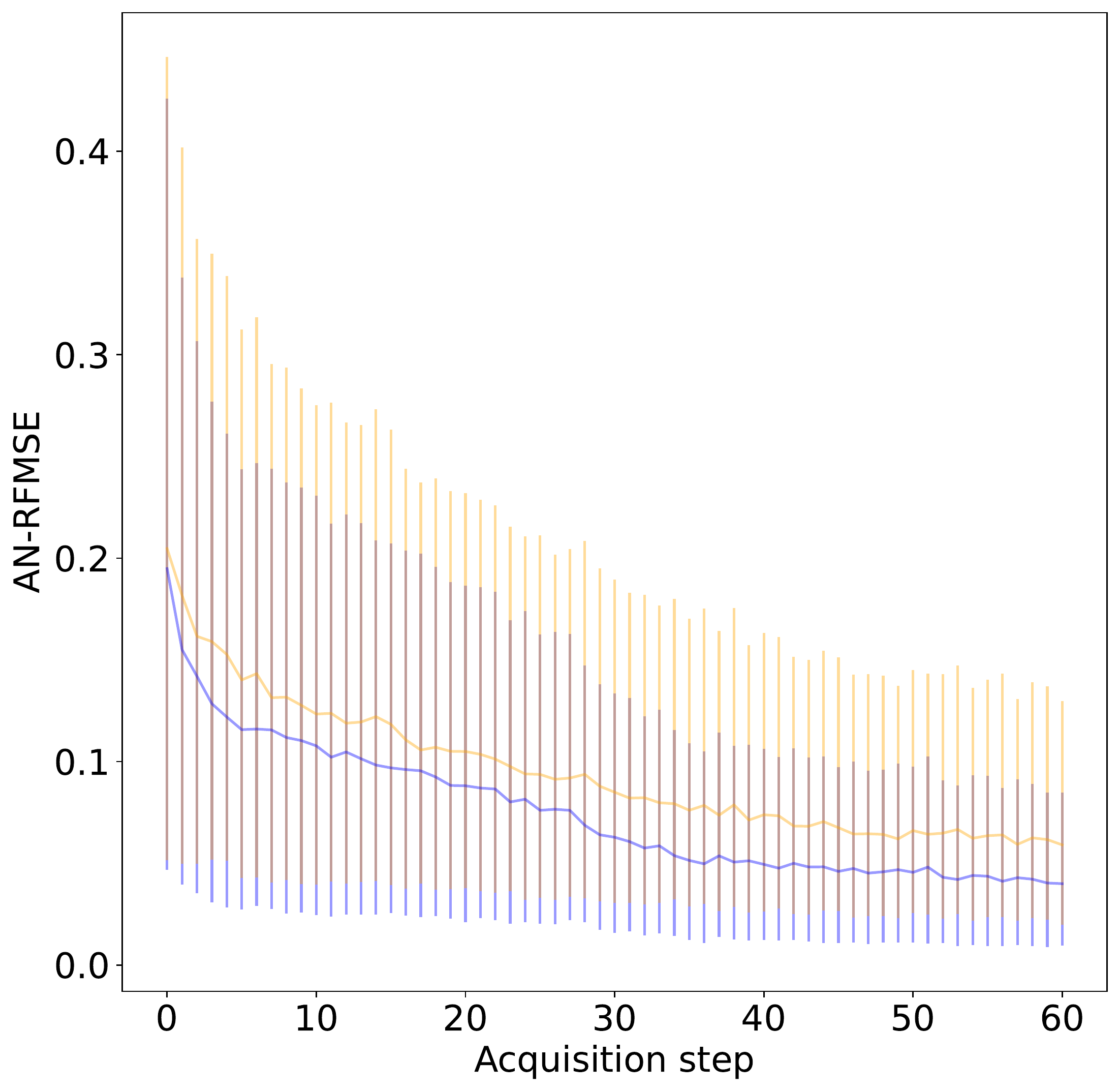}}
        \caption{GI-LI vs. GR-LR.}
        \label{subfig:DMBAL_beta3_vs_RandAprbs}
    \end{subfigure}
    \begin{subfigure}[t]{0.45\linewidth}
        \raisebox{-\height}{\includegraphics[width=\linewidth]{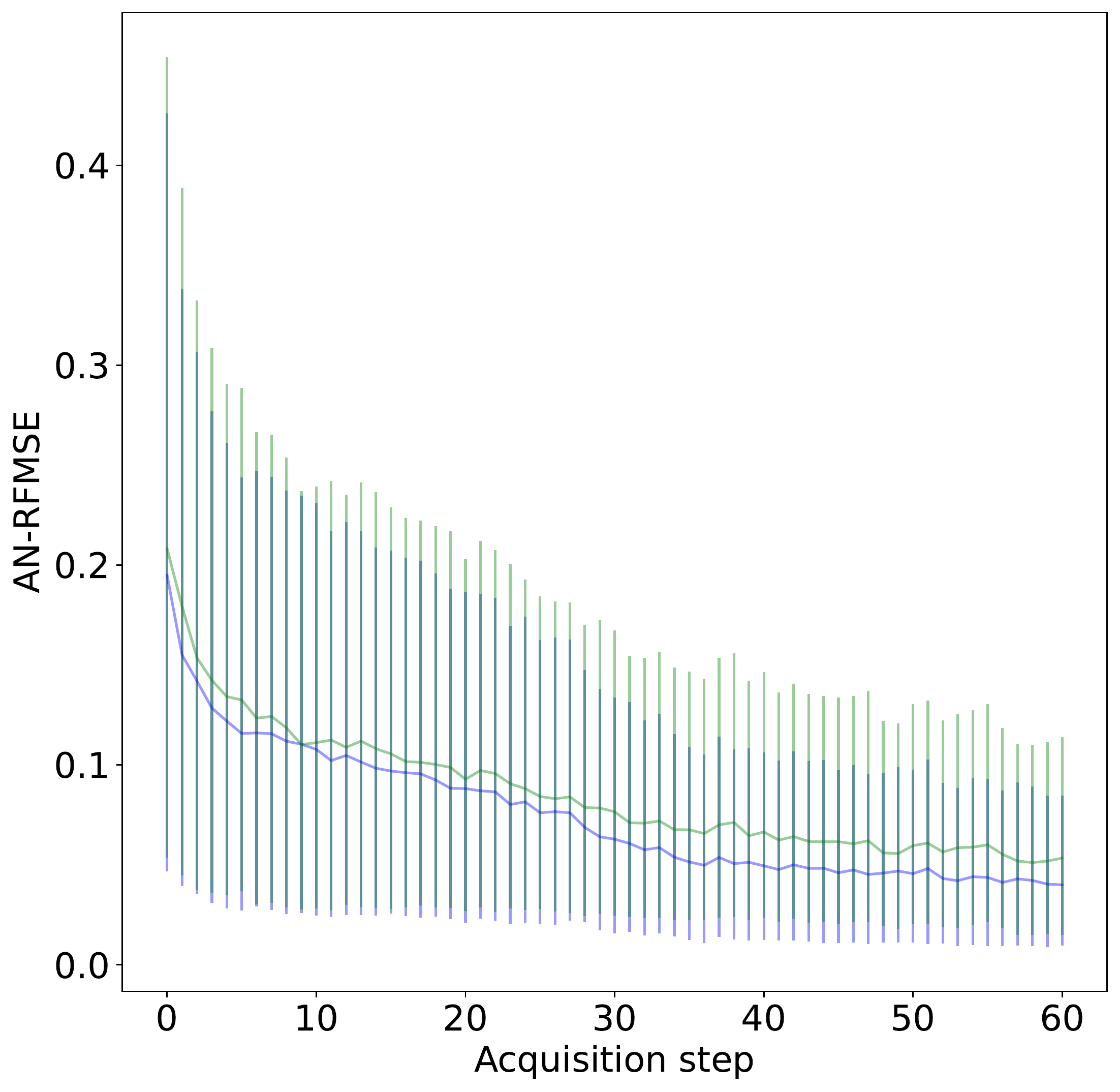}}
        \caption{GI-LI vs. GR-LI.}
        \label{subfig:DMBAL_beta3_vs_RandEnt}
    \end{subfigure}
    
    \begin{subfigure}[t]{0.45\linewidth}
        \raisebox{-\height}{\includegraphics[width=\linewidth]{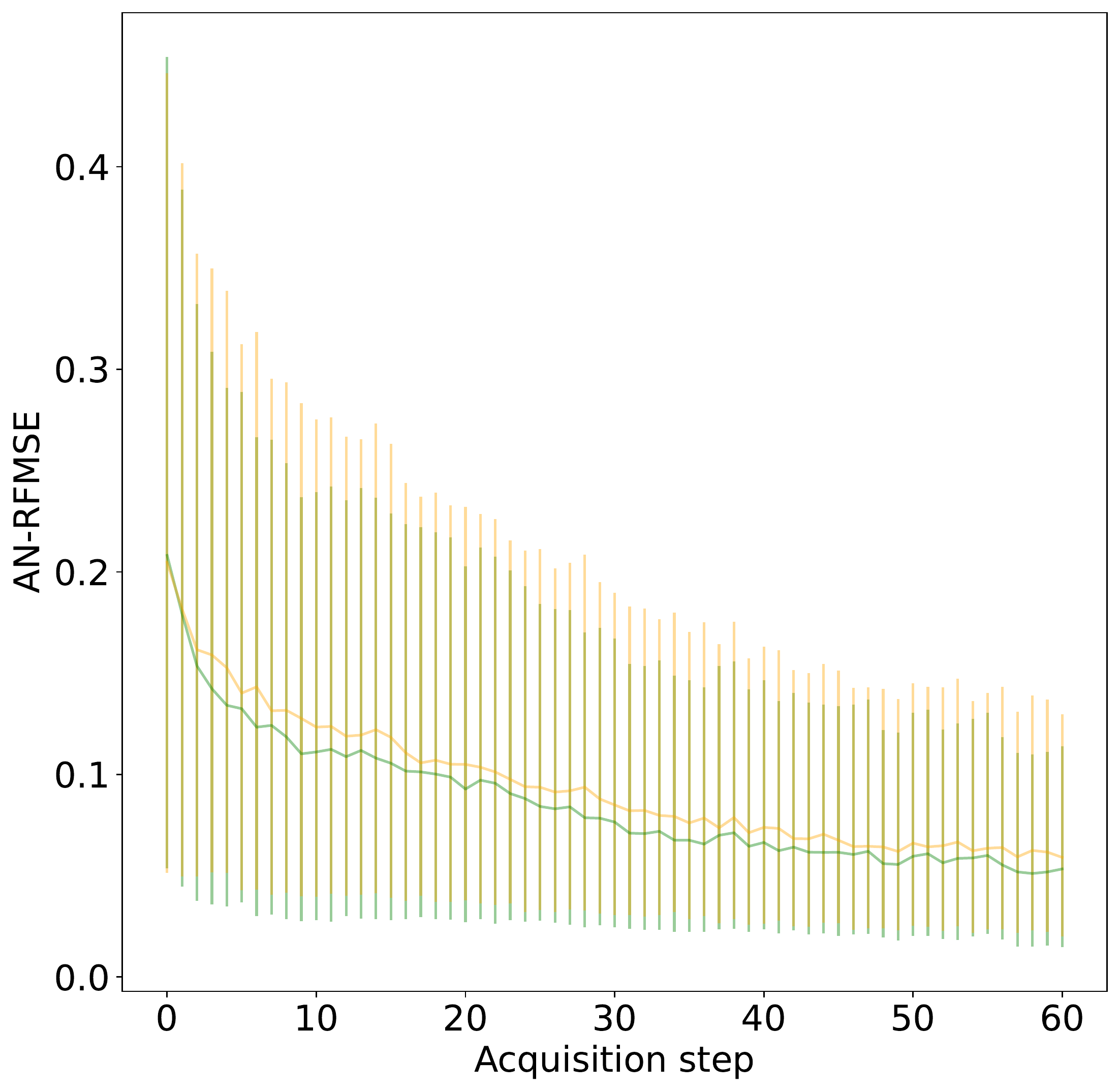}}
        \caption{GR-LI vs. GR-LR.}
        \label{subfig:RandEnt_vs_RandAprbs}
    \end{subfigure}
    \begin{subfigure}[t]{0.45\linewidth}
        \raisebox{-\height}{\includegraphics[width=\linewidth]{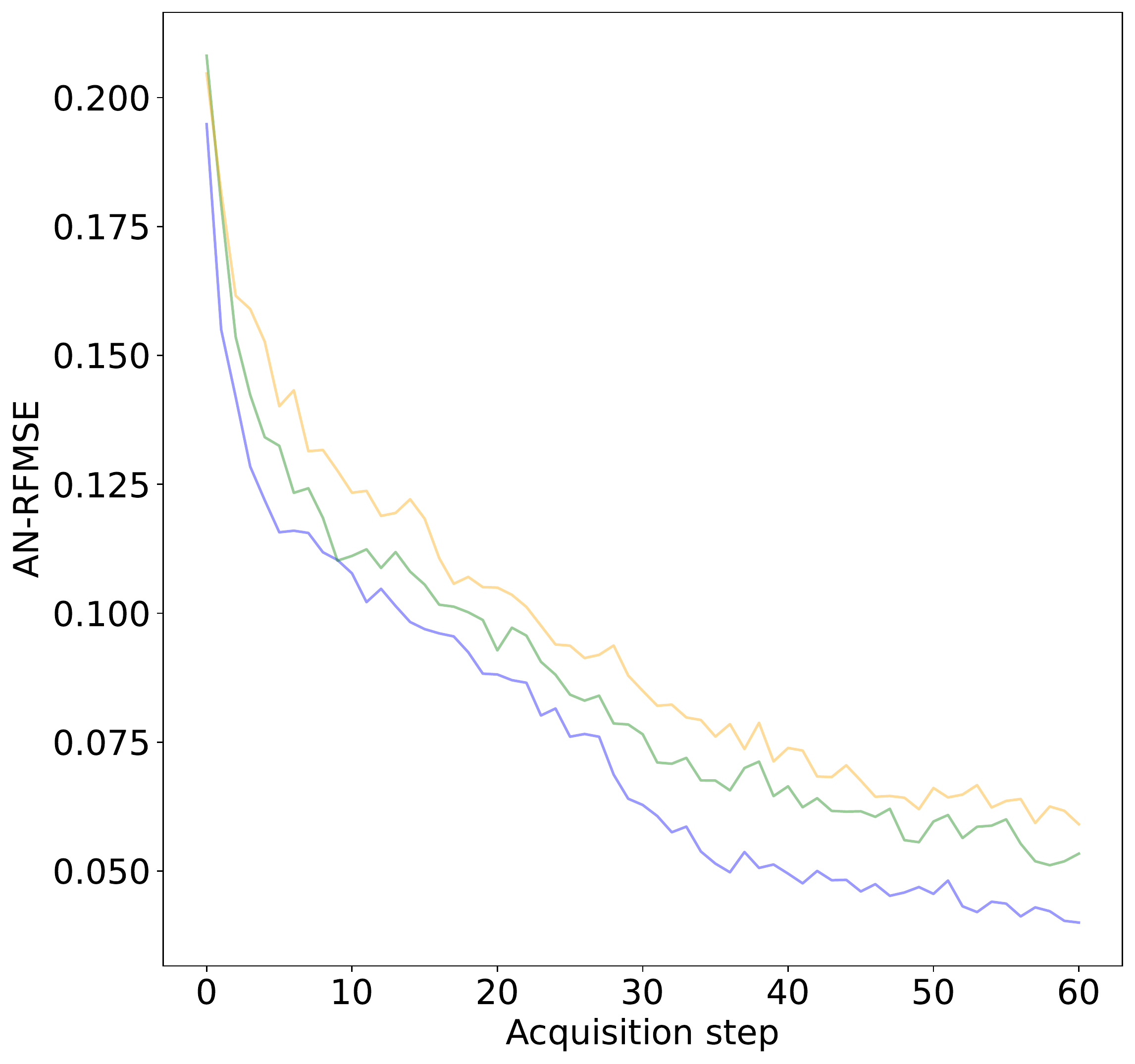}}
        \caption{Mean values GI-LI, GR-LI and GR-LR.}
        \label{subfig:DMBAL_beta3_vs_RandEnt_vs_RandAprbs}
    \end{subfigure}
    
        \begin{adjustbox}{max width=1\linewidth}
        \begin{tikzpicture}
            \begin{customlegend}[legend columns=6,legend style={draw=none, column sep=2ex},legend entries={GI-LI,GR-LI, GRL-R}] 
                \addlegendimage{Tblue!40, fill=Tblue!20, area legend}
                \addlegendimage{Tgreen!40, fill=Tgreen!20, area legend}
                \addlegendimage{Torange!40, fill=Torange!20, area legend}   
            \end{customlegend}
        \end{tikzpicture}
    \end{adjustbox}
    \caption{AN-RFMSE values for NN models trained on each batch of sampled data. The rolling forecast has a prediction horizon of 50 timesteps. The drawn line show the mean AN-RFMSE values, and the error bound show the $25$ prosentile and $75$ procentile of AN-RFMSE values for models trained on data sampled up until the AL loop specified on the x-axis. The GI-LI methods yields significantly lower mean and $75$ procentile of AN-RFMSE values than the two competing methods. The GR-LI method slightly outperforms the GR-LR method, but this result is not as significant.}
    \label{fig:Results_AL_scheemes_vs_Rand}
\end{figure}
Fig.~\ref{fig:Results_AL_scheemes_vs_Rand} presents the performance of the three sampling schemes for after all data acquisition steps. The error bound show the $25 - 75$ percentile of AN-RFMSE values. The upper bound is particularly interesting since it gives an intuition about the models ability to generalize to a broader set of the test set trajectories. The results show that the GI-LI method outperforms the GR-LI and the GR-LR methods, both in terms of higher mean accuracy as well as significantly lower values for the $75$ percentile of AN-RFMSE values. Moreover, the globally random, locally informative GR-LI method shows better performance in terms of higher mean accuracy an lower $75$ percentile AN-RFMSE values than the purely random $GR-LR$ method. However, the superiority of GR-LI method over the GR-LR method is not as significant as the superiority of GI-LI over the two others, indicating that the globally informative step is of major importance.

\begin{figure}[H]
     \centering
    \begin{subfigure}[t]{0.45\linewidth}
        \raisebox{-\height}{\includegraphics[width=\linewidth]{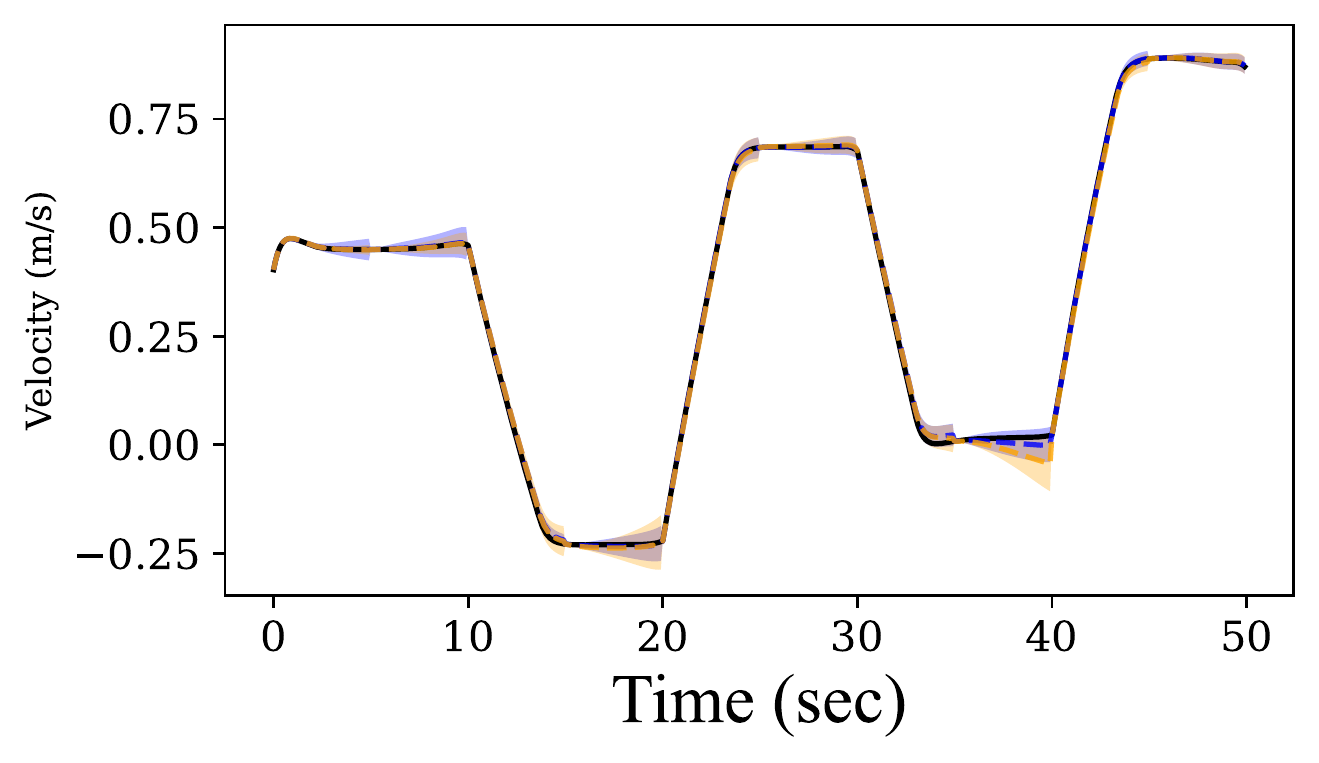}}
        \caption{Surge $x_1$ }
        \label{subfig:AL_timeseries_comp_x1}
    \end{subfigure}
    \begin{subfigure}[t]{0.45\linewidth}
        \raisebox{-\height}{\includegraphics[width=\linewidth]{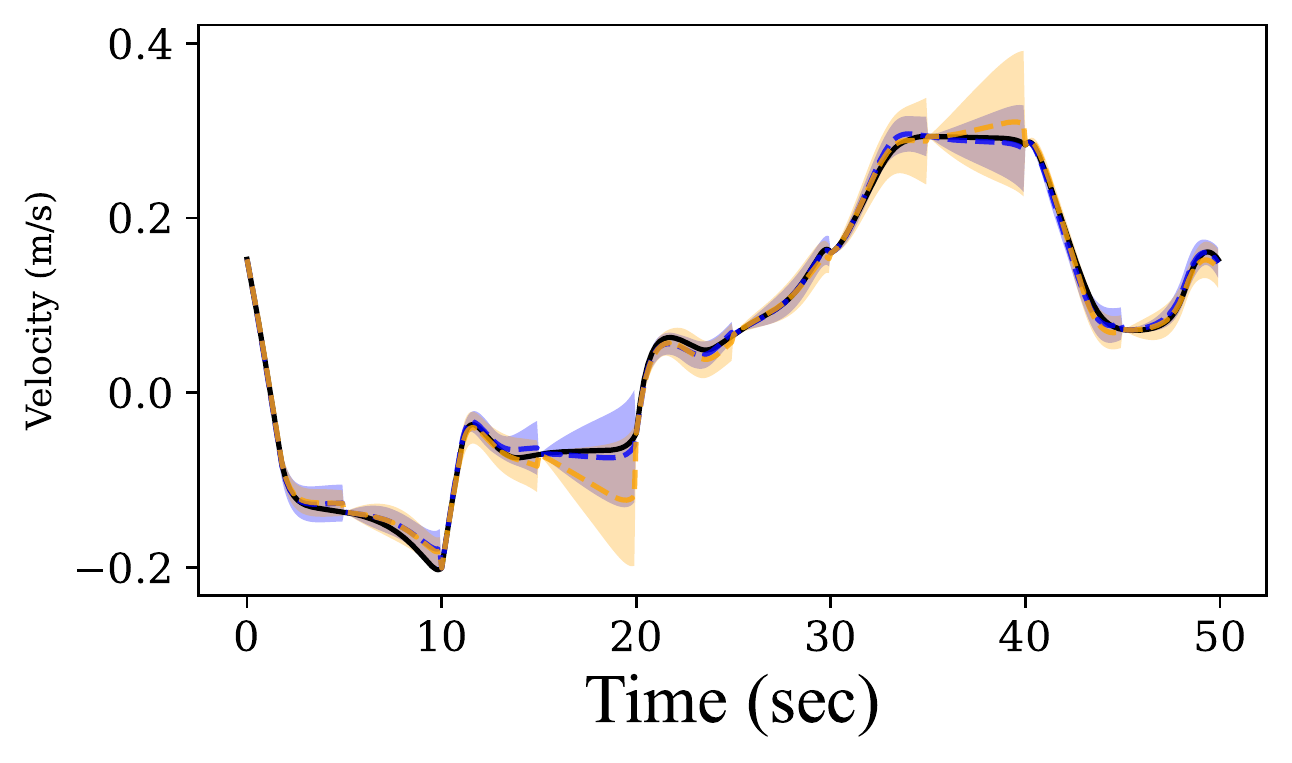}}
        \caption{Sway $x_2$}
        \label{subfig:AL_timeseries_comp_x2}
    \end{subfigure}
    \begin{subfigure}[t]{0.45\linewidth}
        \raisebox{-\height}{\includegraphics[width=\linewidth]{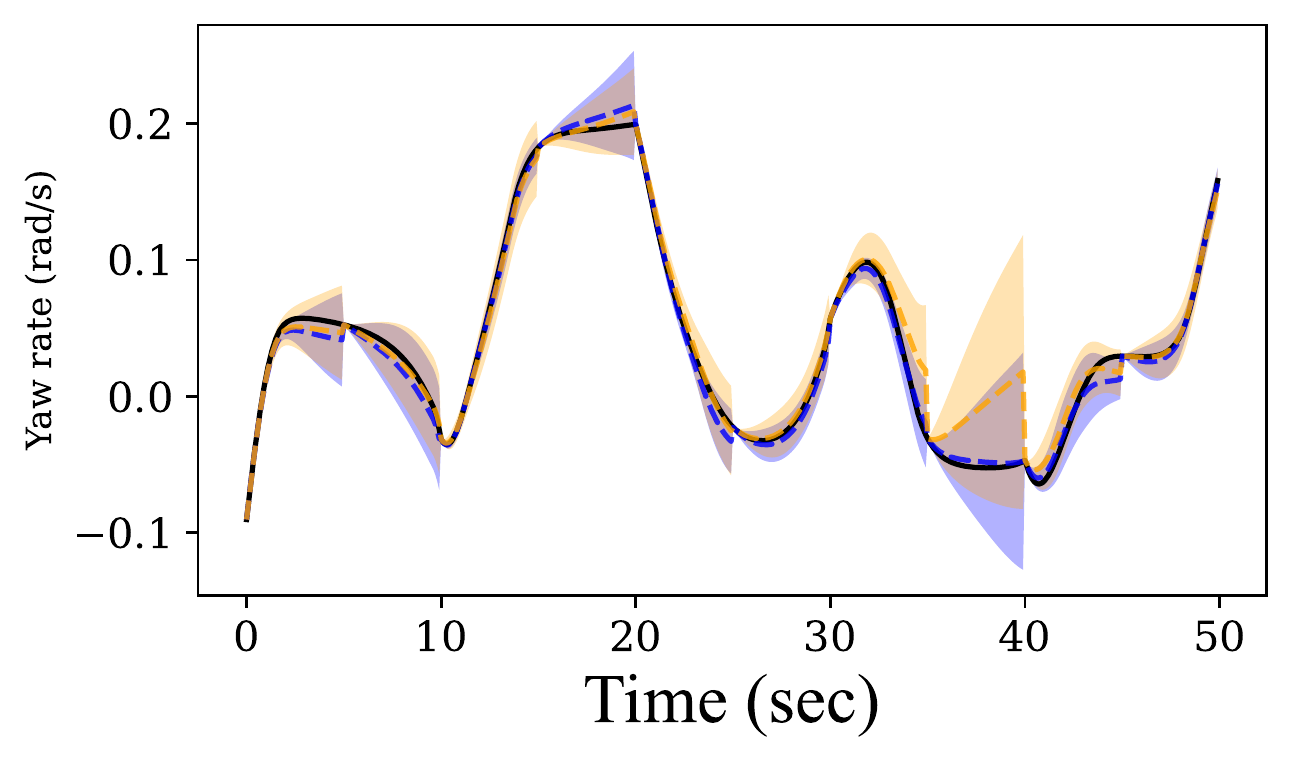}}
        \caption{Yaw $x_3$}
        \label{subfig:AL_timeseries_comp_x3}
    \end{subfigure}
    \begin{subfigure}[t]{0.45\linewidth}
        \raisebox{-\height}{\includegraphics[width=\linewidth]{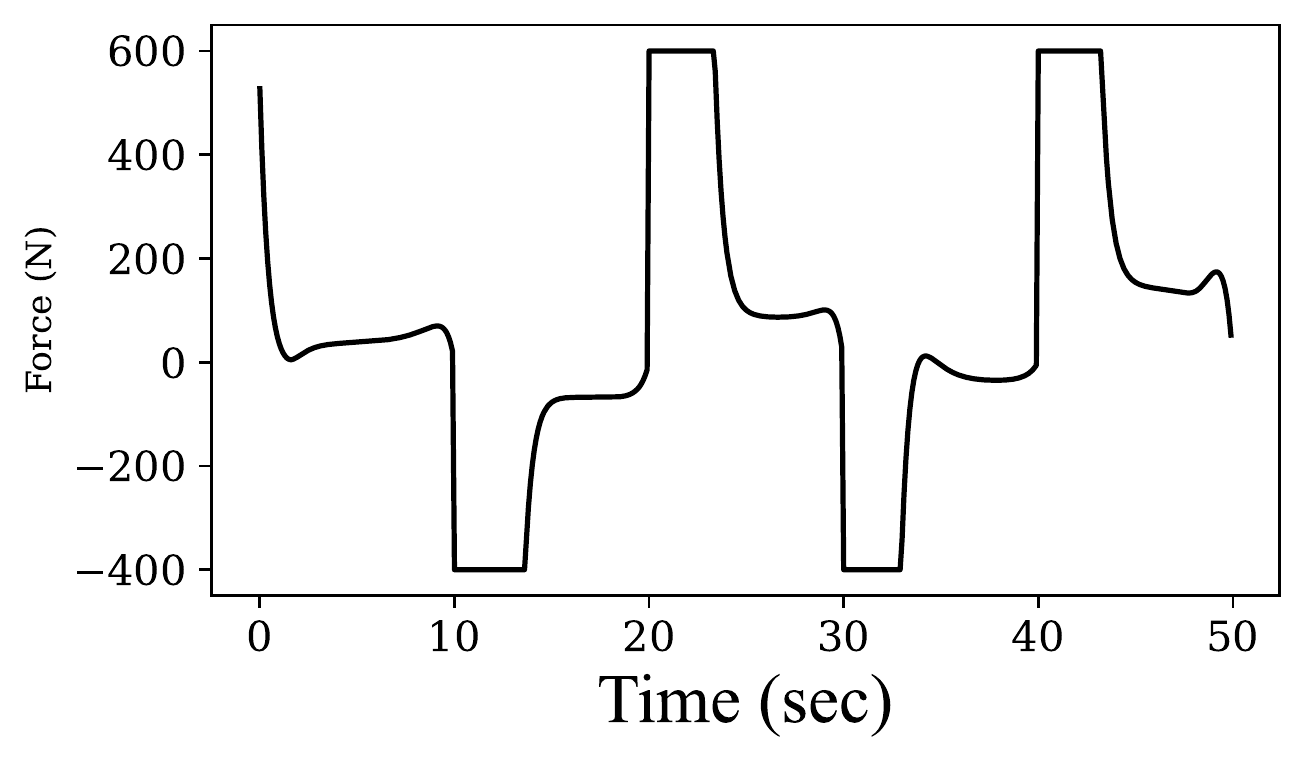}}
        \caption{Forward force $u_1$}
        \label{subfig:AL_timeseries_comp_u1}
    \end{subfigure}
    \begin{subfigure}[t]{0.45\linewidth}
        \raisebox{-\height}{\includegraphics[width=\linewidth]{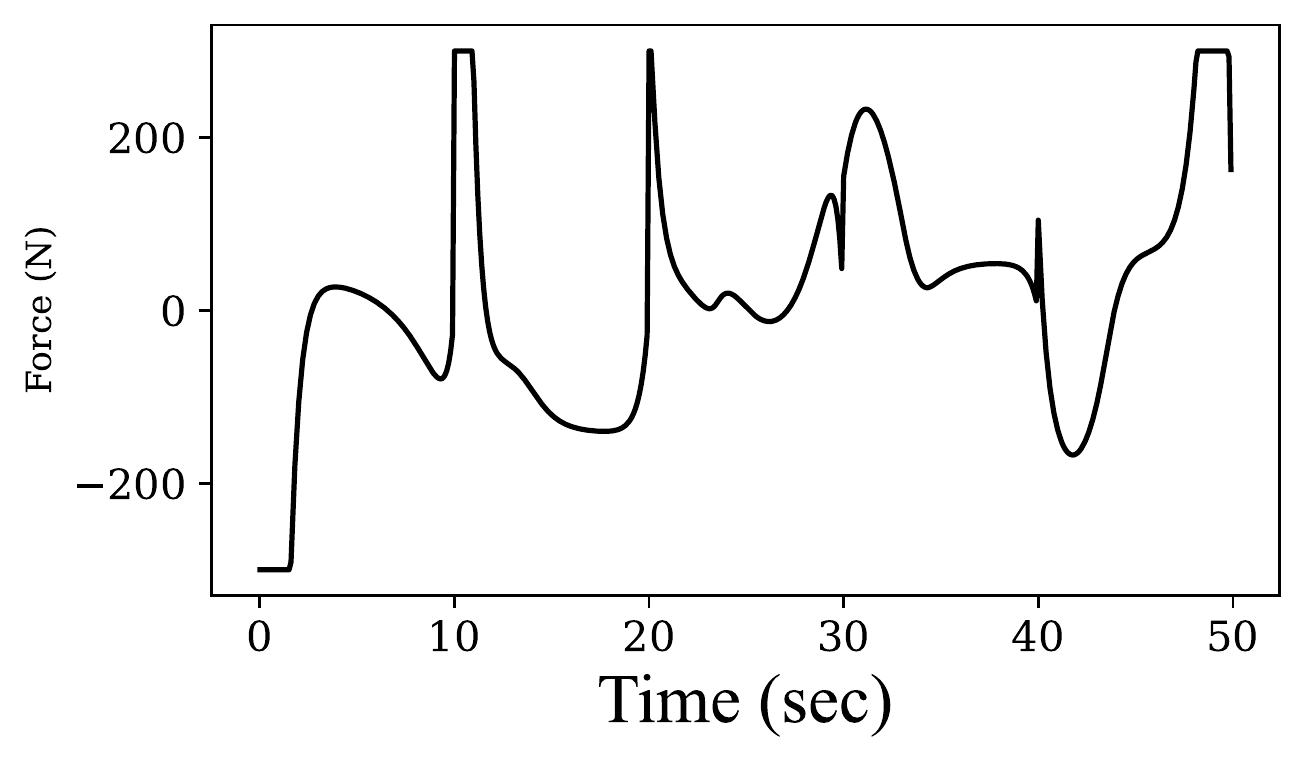}}
        \caption{Sideways force $u_2$}
        \label{subfig:AL_timeseries_comp_u2}
    \end{subfigure}
    \begin{subfigure}[t]{0.45\linewidth}
        \raisebox{-\height}{\includegraphics[width=\linewidth]{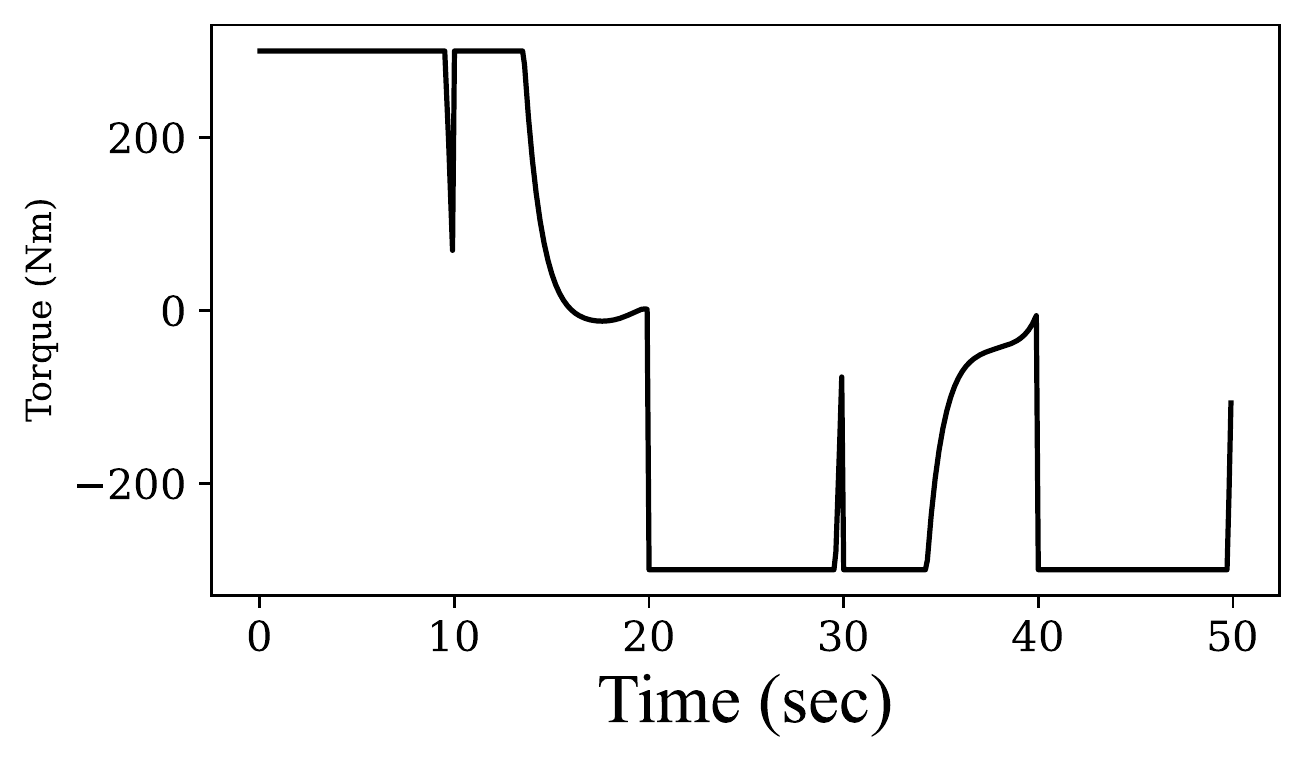}}
        \caption{Torque $u_3$}
        \label{subfig:AL_timeseries_comp_u3}
    \end{subfigure}
    
        \begin{adjustbox}{max width=1\linewidth}
        \begin{tikzpicture}
            \begin{customlegend}[legend columns=6,legend style={draw=none, column sep=2ex},legend entries={GI-LI, GR-LR}] 
                \addlegendimage{Tblue!40, fill=Tblue!20, area legend}
                \addlegendimage{Torange!40, fill=Torange!20, area legend}   
            \end{customlegend}
        \end{tikzpicture}
    \end{adjustbox}
    \caption{Rolling forecast of 50 timesteps. The black drawn lines in Fig.~\ref{subfig:AL_timeseries_comp_x1}- \ref{subfig:AL_timeseries_comp_x1} are state simulated states. The dotted lines show mean forecast values, and the uncertainty bounds show $99.7\%$ confidence intervals. The black lines in Fig.~\ref{subfig:AL_timeseries_comp_u1}-\ref{subfig:AL_timeseries_comp_u3} are control inputs. The ensembles that forecast the states are trained on data based on the GI-LI and GR-LR methods. The plot illustrates how the GI-LI method gives better mean predictions as well as narrower and better calibrated uncertainty bounds.}
    \label{fig:AL_timeseries_comparison}
\end{figure}
Fig.~\ref{fig:AL_timeseries_comparison} shows the mean and uncertainty bounds of ensemble forecasts trained on data sampled with GI-LI and GR-LR methods. The plots illustrate how the GI-LI might provide data that gives improved mean predictions as well as narrower and better calibrated uncertainty bounds.

\subsection{Uncertainty based and hybrid global strategy}
\begin{figure}[H]
    \centering
    \begin{subfigure}[t]{0.45\linewidth}
        \raisebox{-\height}{\includegraphics[width=\linewidth]{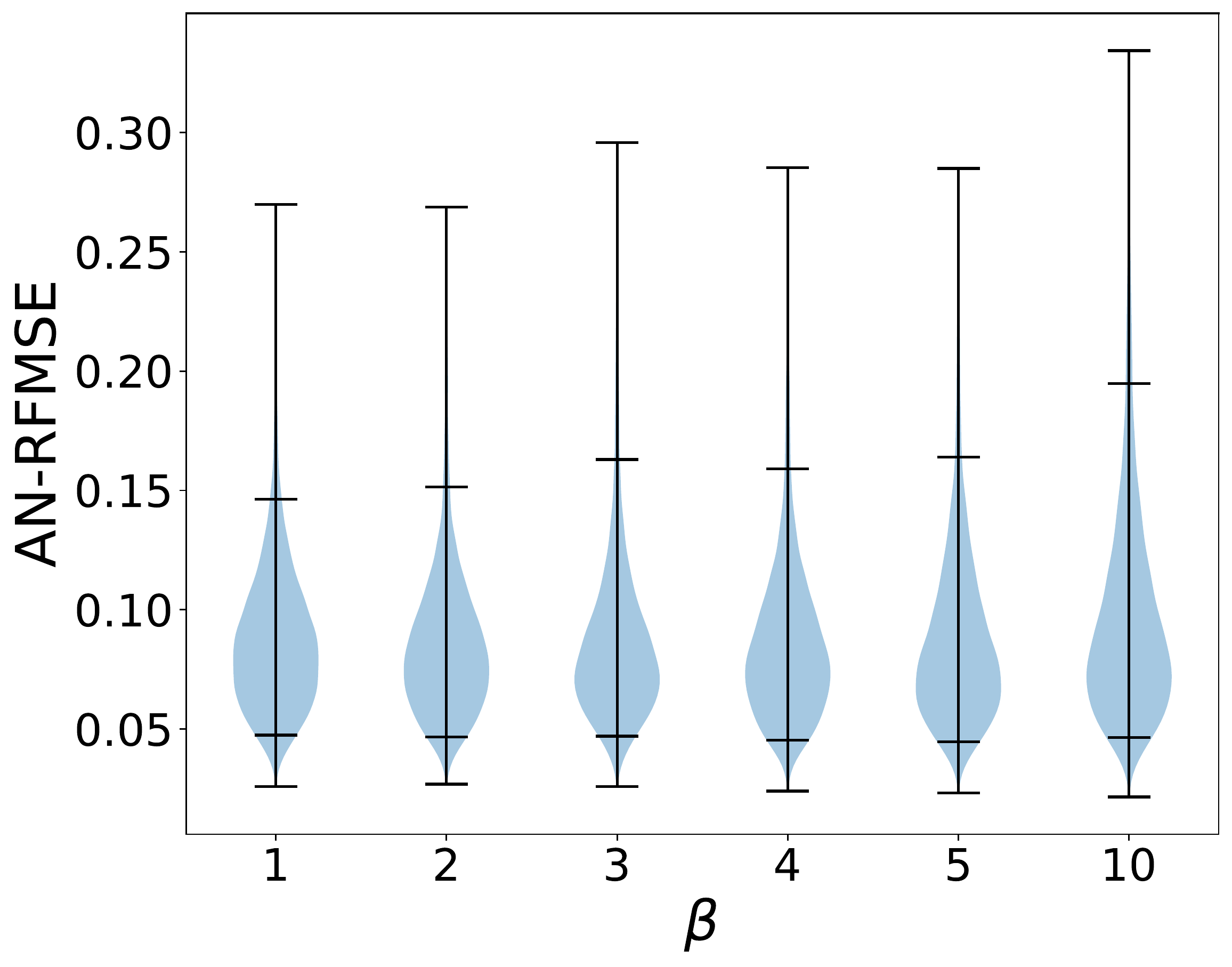}}
        \caption{AN-RFMSE values for models trained between AL loop number ten and 30.}
        \label{subfig:DMBAL_comparison_loop_10_to_30}
    \end{subfigure}
    \begin{subfigure}[t]{0.45\linewidth}
        \raisebox{-\height}{\includegraphics[width=\linewidth]{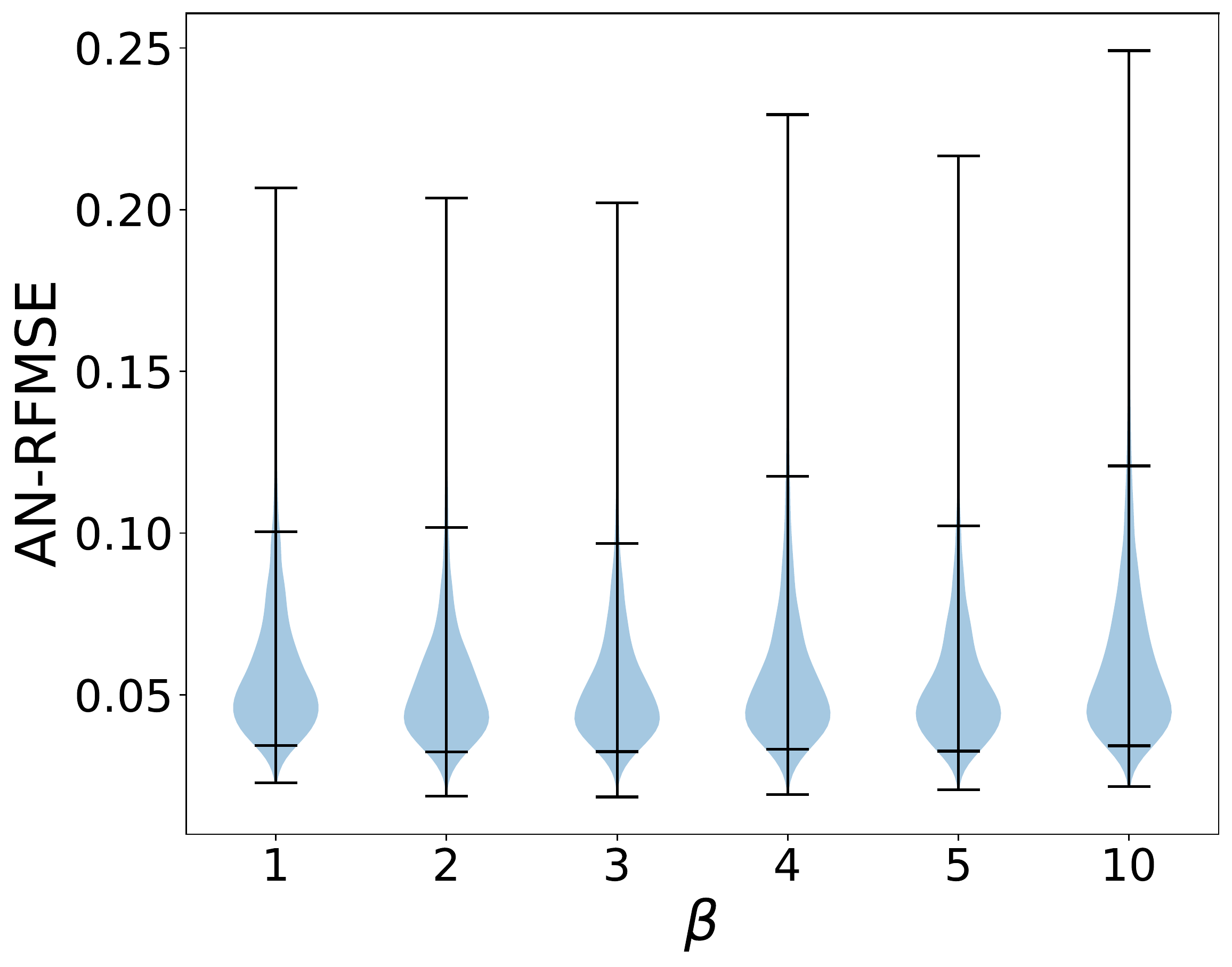}}
        \caption{AN-RFMSE values for models trained between AL loop number 30 and 40.}
        \label{subfig:DMBAL_comparison_loop_30_to_40}
    \end{subfigure}
    \begin{subfigure}[t]{0.45\linewidth}
        \raisebox{-\height}{\includegraphics[width=\linewidth]{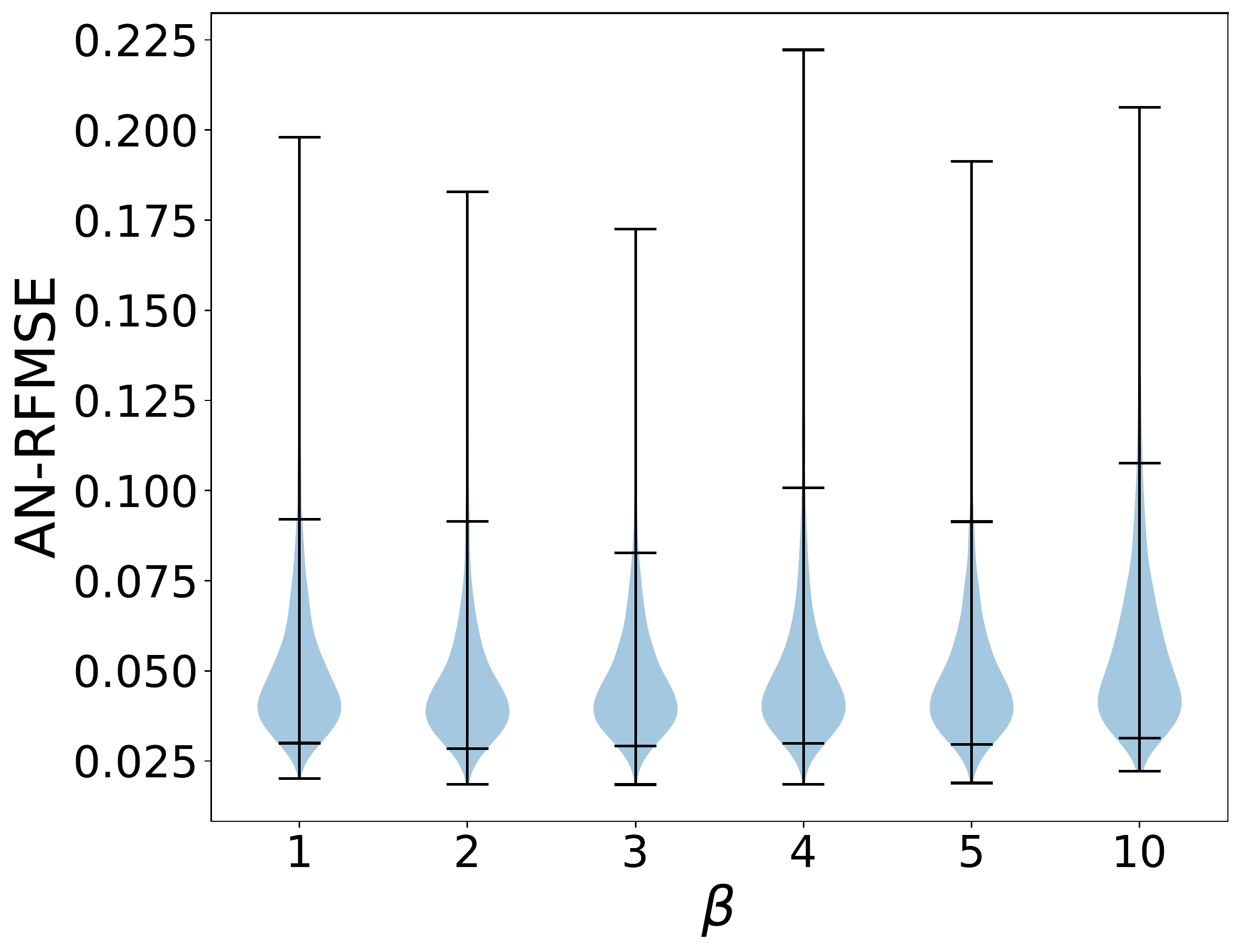}}
        \caption{AN-RFMSE values for models trained between AL loop number 40 and 45.}
        \label{subfig:DMBAL_comparison_loop_40_45}
    \end{subfigure}
    \begin{subfigure}[t]{0.45\linewidth}
        \raisebox{-\height}{\includegraphics[width=\linewidth]{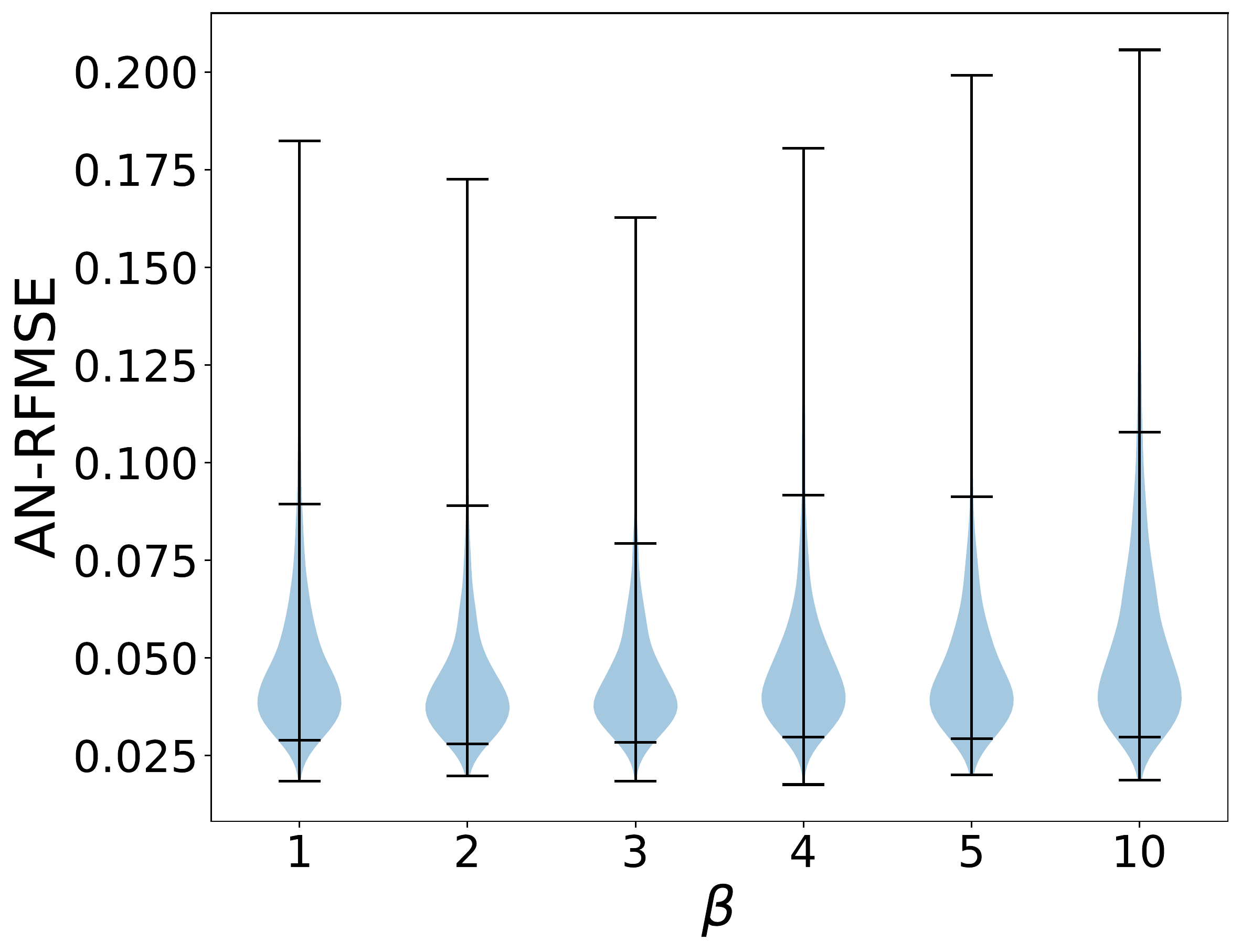}}
        \caption{AN-RFMSE values for models trained between AL loop number 45 and 50.}
        \label{subfig:DMBAL_comparison_loop_45_50}
    \end{subfigure}
    \begin{subfigure}[t]{0.45\linewidth}
        \raisebox{-\height}{\includegraphics[width=\linewidth]{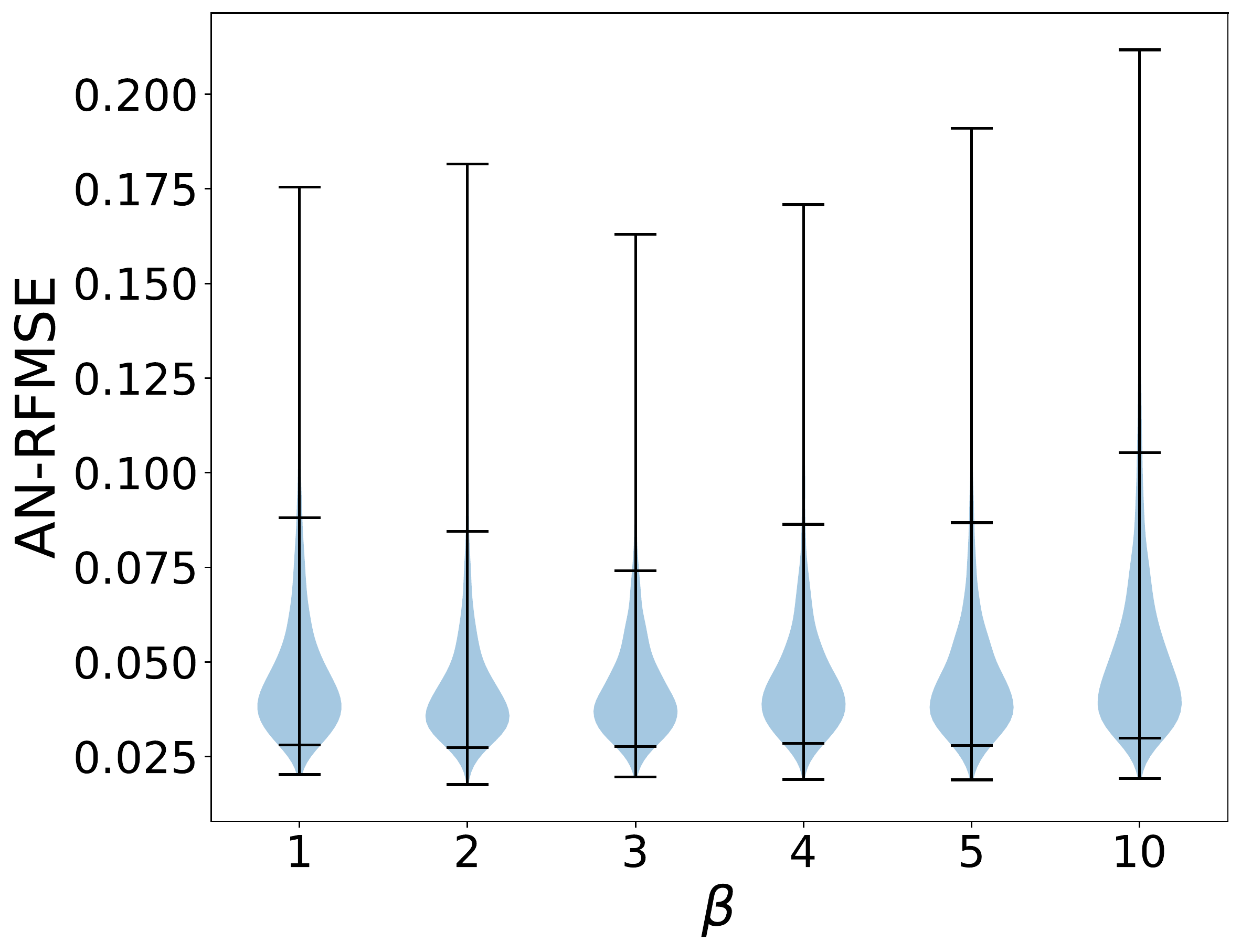}}
        \caption{AN-RFMSE values for models trained between AL loop number 50 and 55.}
        \label{subfig:DMBAL_comparison_loop_50_55}
    \end{subfigure}
    \begin{subfigure}[t]{0.45\linewidth}
        \raisebox{-\height}{\includegraphics[width=\linewidth]{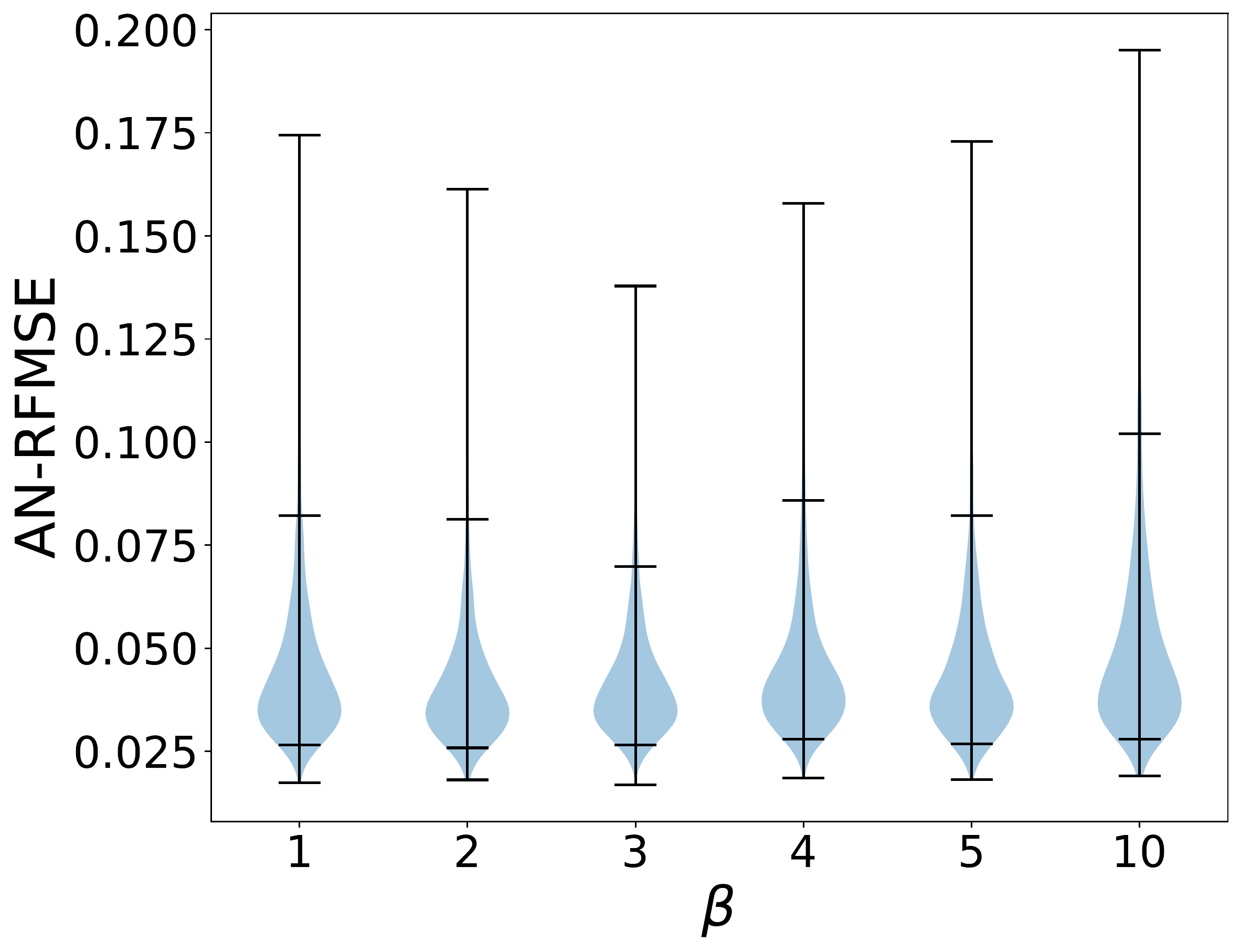}}
        \caption{AN-RFMSE values for models trained between AL loop number 55 and 60.}
        \label{subfig:DMBAL_comparison_loop_55_60}
    \end{subfigure}
    \caption{Violin plots of AN-RFMSE values corresponding to models trained on data sampled with the GI-LI method. The rolling forecast has a prediction horizon of 50 timesteps. All methods use the global DMBAL method for choosing initial conditions, but with different prefilter hyperparameter $\beta$. The two innermost horizontal lines of each violin plot show the $5$ and $95$ procentile, while the outermost horizontal lines show the extreme values. Each subplot includes AN-RFMSE values for models trained on data from a range of the AL loops. That is,  Fig.~\ref{subfig:DMBAL_comparison_loop_10_to_30} summarize the perfomance of the sampling methods in early stages, while Fig.~\ref{subfig:DMBAL_comparison_loop_40_45}-\ref{subfig:DMBAL_comparison_loop_55_60} summarize the performace later in the AL scheme.}
    \label{fig:violin_plot_AL_DMBAL_beta_comp}
\end{figure}
Fig.~\ref{fig:violin_plot_AL_DMBAL_beta_comp} show the performance of GI-LI methods using the global DMBAL method with different prefilter parameter $\beta$ for different stages of the AL scheme. The algorithm only considers the $b \cdot \beta$ samples with highest informativeness score according to a given informativeness measure, where a batch $\mathcal{B}^*$ of $b=10$ initial conditions are chosen. Hence, $\beta=1$ corresponds to simply picking the $b$ samples with the highest informativeness score. Since the chosen informative measure is uncertainty based, using $\beta=1$ means that the global method is uncertainty based. Using $\beta>1$ means that the global algorithm takes into account diversity by choosing the samples closest to centroid centers of a k-means algorithm, which means that the overall method is a hybrid AL strategy, combining uncertainty measures and diversity measures. Fig~\ref{fig:violin_plot_AL_DMBAL_beta_comp} shows that the average performance in all plots is approximately the same for all choices of $\beta$. In Fig.~\ref{subfig:DMBAL_comparison_loop_10_to_30} and \ref{subfig:DMBAL_comparison_loop_10_to_30} showing performance for models trained on data acquired after respectively $10$ to $30$ AL loops and $30$ to $40$ AL loops, it is difficult to conclude any significant differences, other than that the method using $\beta=10$ has higher extremum AN-RFMSE values, as well as more stretched distribution towards higher AN-RFMSE values, indicating that lower choices of $\beta$ give better results. Fig~\ref{subfig:DMBAL_comparison_loop_40_45}-\ref{subfig:DMBAL_comparison_loop_55_60} show the performance of the method for different $\beta$ values for later stages of the AL scheme. The most evident result is the tendency that the upper extreme values of AN-RFMSE have a minimum for $\beta=3$, and that both lower and higher values of $\beta$ give higher extreme AN-RFMSE values. Moreover, it seems like the $95$ percentile also is the lowest for $\beta=3$. Moreover, again, using high values of $\beta$ turns out to give worse result as $\beta=10$ show the worst performance. The overall results from the conducted case study indicate that using the DMBAL algorithm in the global acquisition scheme is of minor importance compared to acquiring a batch of samples with the highest uncertainty scores. However, choosing the right values of the prefilter hyperparameter $\beta$, the modeling errors in extreme cases can potentially be reduced. The limited value of the hybrid method used in this study might be explained by that it only compares the similarity of initial states rather than similarities between candidate trajectories. Moreover, the shortcomings of uncertainty based methods are known to be more present if the batch size is greater. With the currently choice of batch size $b=10$ trajectories being sampled at each acquisition step, the power of hybrid strategies might not be present.
\section{Conclusions and future work}
\label{sec:conclusions}
The main conclusions from the work can be itemized as follows:
\begin{itemize}
    \item The globally random, locally informative based GR-LI strategy show slightly better results than a globally random, locally random strategy in terms of mean accuracy an lower values of the $75$ procentile of AN-RFMSE values, indicating better generalization. However, the novel GI-LI DeepAL scheme significantly outperform GR-LI and GR-LR schemes both in terms of mean accuracy and $75$ procentile values of AN-RFMSE. This indicates that global explorations are of major importance with respect to achieving higher accuracy and generalization.

    \item The DMBAL  approach, which emphasizes diversity in the selection of samples, might reduce the upper bound of extreme error values, provided that the prefilter hyperparameter is carefully chosen. This method is compared to simply selecting the top $b$ samples based on some uncertainty measure, when sampling a batch of initial conditions globally. However, the DMBAL approach does not exhibit significant improvements over the global uncertainty based method beyond this in the given case study. The method only compares the similarity between initial states of the candidate trajectories, rather than similarity between the whole candidate trajectories. This leaves out potentially important information. 
\end{itemize}
The novel DeepAL framework is flexible and allows for a range of AL aqcuisition strategies. Conducting a comparative study including different AL acquisition functions in the framework would be highly interesting as it could increase our knowledge about efficient sampling of dynamical systems. Global hybrid strategies that can consider similarity of entire state trajectories are of particular interest since the currently tested hybrid strategy that only compares initial conditions seems to be of limited value. 

\section{Acknowledgments}
This work was supported by the industry partners Borregaard, Elkem, Hydro, Yara and the Research Council of Norway through the project TAPI: Towards Autonomy in Process Industries, project number 294544.

\bibliographystyle{elsarticle-num} 
\bibliography{ref.bib}

\clearpage
\newpage
\appendix

\end{document}